\documentclass{article}

\usepackage{longcat_style}
\usepackage{adjustbox}
\usepackage[utf8]{inputenc}
\usepackage[T1]{fontenc}
\usepackage{CJKutf8}
\usepackage{newunicodechar}
\usepackage{hyperref}
\usepackage{xcolor}
\usepackage[normalem]{ulem}
\hypersetup{
  colorlinks=true,
  linkcolor=blue,
  urlcolor=blue,
  citecolor=blue,
  linkbordercolor=blue,
  urlbordercolor=blue,
  citebordercolor=blue,
  pdfborderstyle={/S/U/W 1},
}

\usepackage{float}
\usepackage{placeins}

\usepackage{url}
\usepackage{booktabs}
\usepackage{amsfonts}
\usepackage{amsmath}
\usepackage{amssymb}
\usepackage{amsthm}
\usepackage{mathtools}
\usepackage{nicefrac}
\usepackage{microtype}
\usepackage{graphicx}
\usepackage{natbib}
\usepackage{doi}
\usepackage{xspace}
\usepackage{enumitem}
\usepackage{multirow}
\usepackage{array}
\usepackage{subcaption}
\usepackage{makecell}
\usepackage{cleveref}
\usepackage{pifont}
\usepackage[most]{tcolorbox}
\usepackage{siunitx}
\usepackage{tikz}
\usepackage{pgfplots}
\pgfplotsset{compat=1.16}
\usetikzlibrary{positioning,arrows.meta,fit,backgrounds,calc,decorations.pathreplacing}

\setlist[itemize]{leftmargin=*}
\setlist[enumerate]{leftmargin=*}
\setlist[description]{leftmargin=*}

\newcommand{\gmoe}{\textsc{GMoE}\xspace}

\newcommand{\concurrent}{MH-LatentMoE\xspace}

\newcommand{\R}{\mathbb{R}}
\newcommand{\ngroup}{G}
\newcommand{\dsub}{d_{\mathrm{sub}}}
\newcommand{\Nexp}{N'}
\newcommand{\kexp}{k'}

\title{Grouped Mixture-of-Experts}

\author{%
\normalfont Chao Wang\textsuperscript{*}, Tao Zhou\textsuperscript{*}, Jiaqi Zhang\textsuperscript{\dag}, Han Lu, Yuwei Jiang, Mingliang Zhang\\
Hong Liu, Chang Lu, Weilin Cai, Gang Liu, Bole Zhou, Tao Li, Ge Li\\
Yipeng Mei, Yifan Lu, Xue Wang, Peng Pei, Yuchen Xie, Xunliang Cai\\[4pt]
\textbf{Meituan LongCat Team}%
}

\renewcommand{\headeright}{\raisebox{-0.2\height}{\includegraphics[height=1.8em]{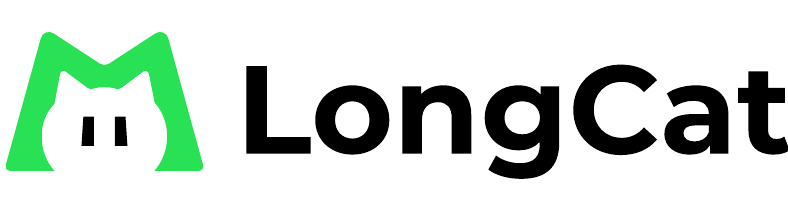}}}
\renewcommand{\shorttitle}{Grouped Mixture-of-Experts}

\begin{document}
\maketitle
\makeatletter
\begingroup
\renewcommand{\thefootnote}{}
\long\def\@makefntext#1{\noindent#1}
\footnotetext[0]{\textsuperscript{*}Equal contribution.\quad \textsuperscript{\dag}Corresponding author: \href{mailto:zhangjiaqi39@meituan.com}{zhangjiaqi39@meituan.com}.}
\endgroup
\makeatother

\begin{abstract}
Fine-grained expert segmentation has become a cornerstone of modern mixture-of-experts (MoE) architectures. 
Extending this design along an orthogonal dimension, we introduce Grouped Mixture-of-Experts (\gmoe),
which projects the hidden state into multiple low-dimensional groups, each with an independent router and dedicated expert pool, before merging their outputs back into the residual stream. 
To integrate this grouped architecture into established fine-grained MoE training recipes, we develop a training-aligned parameterization that enables their reuse without additional hyperparameter tuning.
Together, the architecture and its tailored parameterization improve sparsity scaling at a constant active-compute budget. Power-law fits indicate that fine-grained MoE requires approximately $1.79\times$ \gmoe{}'s sparsity ratio to achieve the same validation loss.
Scaling up to 69B parameters, \gmoe{} outperforms its fine-grained baseline on all 13 benchmarks after 5T training tokens and surpasses LatentMoE on 10 of 13 benchmarks after 1.4T tokens.
At the systems level, placing each group's expert pool on a single rank under expert parallelism keeps dispatch and combine local, eliminating routing-dependent workload imbalance across ranks, while kernel optimizations further accelerate local execution.
Collectively, these results demonstrate that \gmoe improves both model performance and distributed training efficiency.
\end{abstract}

\begingroup
\setlength{\intextsep}{6pt}
\begin{figure}[H]
\centering
\setlength{\abovecaptionskip}{6pt}
\includegraphics[width=0.90\textwidth]{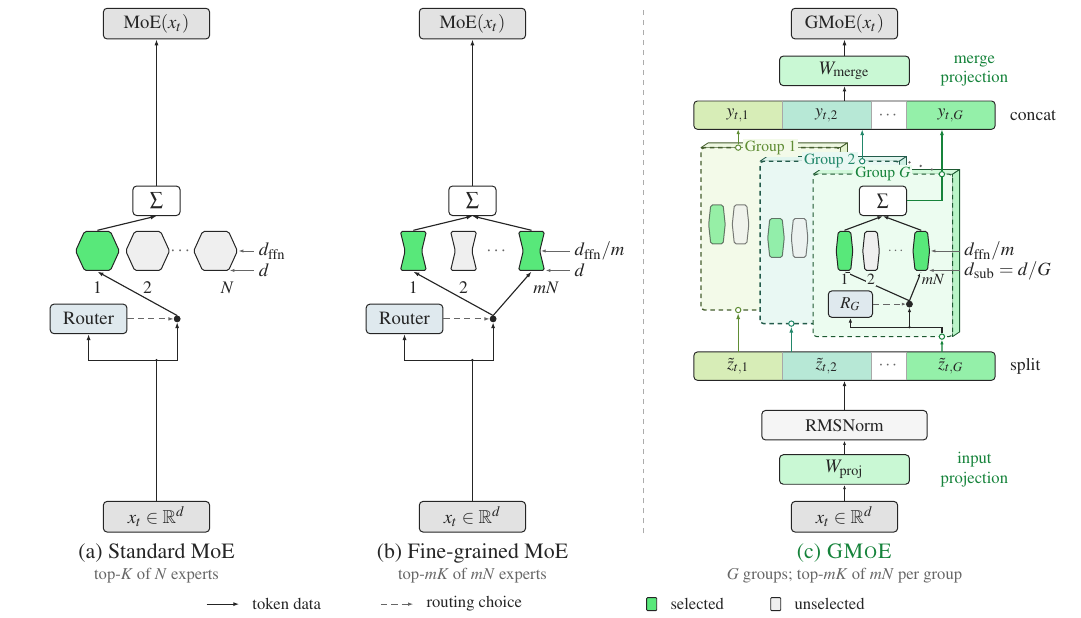}
\caption{\textbf{Architectural comparison.} \textbf{(a) Standard MoE} selects the top-$K$ of $N$ experts of shape $d \rightarrow d_{\mathrm{ffn}} \rightarrow d$. \textbf{(b) Fine-grained MoE}~\citep{dai2024deepseekmoe} partitions the intermediate dimension by $m$, yielding $mN$ experts of shape $d \rightarrow d_{\mathrm{ffn}}/m \rightarrow d$ and activating the top-$mK$ at matched FLOPs. \textbf{(c) \gmoe} projects and normalizes each token with $W_{\mathrm{proj}}$ and RMSNorm, then splits it into $\ngroup$ subspaces of width $d_{\mathrm{sub}}=d/\ngroup$. Each has its own router and experts of shape $d_{\mathrm{sub}} \rightarrow d_{\mathrm{ffn}}/m \rightarrow d_{\mathrm{sub}}$, giving $\ngroup mN$ experts with $\ngroup mK$ activated per token. Group outputs are concatenated and merged by $W_{\mathrm{merge}}$. Only group $G$ is expanded; each $\sum$ denotes routing-weighted aggregation.}
\label{fig:arch}
\end{figure}

\endgroup

\clearpage
\tableofcontents
\clearpage

\section{Introduction}

Scaling remains the most reliable driver of progress in large language models (LLMs).
For dense Transformers, jointly scaling model size and training data leads to consistent performance gains, as demonstrated by the Chinchilla study~\citep{hoffmann2022chinchilla}.
Sparse mixture-of-experts (MoE)~\citep{shazeer2017moe,fedus2022switch,zoph2022stmoe} opens a complementary scaling dimension through sparsity, increasing the total number of experts while activating only a subset for each token.
On top of this standard sparse recipe, DeepSeekMoE~\citep{dai2024deepseekmoe} introduced a further architectural dimension: fine-graining the experts.
Starting from $N$ experts with $K$ activated per token, it partitions each expert's FFN intermediate dimension by a factor $m$, yielding $mN$ smaller experts and activating $mK$ per token.
At a matched routed-expert FFN compute budget, this expands the combinations of activated experts and supports finer specialization.
This fine-grained recipe has since become the dominant design for open-source MoE LLMs.

While fine-graining narrows the FFN intermediate dimension, each expert still processes the full hidden representation. In this report, we introduce Grouped Mixture-of-Experts (\gmoe), which adds hidden-space grouping as an orthogonal architectural scaling dimension.
A \gmoe layer linearly projects each token representation, stabilizes its scale via inner normalization, and splits the projected state into $\ngroup$ equal-width subspaces. Each subspace is processed by an independent fine-grained MoE with its own router and expert pool. Every token is processed across all $\ngroup$ groups, with expert selection performed independently within each group. The group outputs are concatenated and merged back into the residual stream (Figure~\ref{fig:arch}).

Closely related to our work, Multi-Head LatentMoE (\concurrent)~\citep{cui2026mhlatentmoe}, which extends LatentMoE~\citep{elango2026latentmoe}, approaches a structurally similar multi-head latent routing layer from a systems-efficiency perspective, proposing Head Parallel and IO-aware kernels to reduce communication overhead and enable deterministic execution.
Our work instead investigates hidden-space grouping as an orthogonal architectural dimension for fine-grained MoE, introducing training-aligned parameterization and load-balancing calibration to make this dimension effective in practice.
Building on this formulation, we characterize sparsity scaling and the representation dynamics underlying the grouping advantage, and validate these gains through 5T-token pre-training at the 69B-parameter scale (3.8B activated).
On the systems side, we pair group-local expert placement with FP8 kernel fusion and static-buffer execution, evaluating end-to-end training efficiency at the 69B scale.

In summary, our main contributions are threefold:

\begin{itemize}
    \item \textbf{GMoE Architecture and Training Alignment.}
    We introduce \gmoe, which adds hidden-space grouping as an orthogonal architectural dimension. Alongside this design, we develop a training-aligned parameterization and load-balancing calibration to reuse fine-grained MoE training recipes without additional hyperparameter tuning.
    \item \textbf{Scaling Behavior and Modeling Benefits.} 
    We study the sparsity scaling of \gmoe over a $16\times$ sparsity sweep. Power-law fits indicate that fine-grained MoE requires approximately $1.79\times$ \gmoe{}'s sparsity ratio to achieve the same validation loss. Scaling up to 69B parameters, \gmoe outperforms the fine-grained baseline on all 13 benchmarks after 5T tokens and surpasses LatentMoE on 10 of 13 benchmarks after 1.4T tokens. Further representation analysis reveals that, despite operating on a shared hidden state, groups diverge in their routing behavior and contribute non-redundant updates to the residual stream.

    \item \textbf{Systems Optimization for Balanced Large-Scale Training.}
    By aligning expert parallelism with the group layout, we place each group's complete expert pool on a single rank, keeping dispatch and combine local. Each rank receives an equal number of subspace tokens, with fixed-volume cross-rank communication independent of routing decisions. To further reduce local execution overhead, we pair this layout with FP8 kernel fusion and static-buffer execution, enabling efficient end-to-end training at the 69B scale.
\end{itemize}

\section{Related Work}

\paragraph{Fine-grained MoE.}
Sparse MoE architectures increase total model capacity while activating only a subset of experts for each token~\citep{fedus2022switch,lepikhin2021gshard}.
DeepSeekMoE~\citep{dai2024deepseekmoe} increases routing granularity by reducing the expert intermediate width and expanding the expert pool, a design adopted in subsequent models~\citep{deepseekv3,qwen3moe,kimiteam2025kimik2,glmteam2026glm5}.
\gmoe extends fine-graining to the hidden-space dimension, increasing the total number of experts and enabling richer combinations through independent routing across groups.

\paragraph{Latent-Space MoE.}
LatentMoE~\citep{elango2026latentmoe} projects token representations into a shared low-dimensional latent space for routed-expert computation and communication, while retaining routing in the original hidden space.
Building on LatentMoE, \concurrent~\citep{cui2026mhlatentmoe} introduces multiple latent heads with independent routers and expert pools.
\gmoe shares the independently routed subspace structure of \concurrent, with an emphasis on training-aligned parameterization for reusing fine-grained MoE training recipes and on sparsity scaling behavior, complemented by large-scale model validation.

\paragraph{Load Balancing and Distributed Execution.}
Auxiliary load-balancing losses encourage uniform expert utilization in sparse MoE~\citep{fedus2022switch}, while auxiliary-loss-free methods adjust expert selection via dynamically updated routing biases~\citep{wang2024lossfree}.
Kimi K3~\citep{kimiteam2026kimik3} introduces Quantile Balancing, deriving these bias updates from routing-score margin quantiles.
Even with router-level balancing, dynamic selection can still induce transient workload skew and variable traffic across expert-parallel ranks.
At the systems level, Kimi K3's MoonEP addresses this via online planning and dynamic expert replication, whereas \concurrent~\citep{cui2026mhlatentmoe} and \gmoe take a structural approach.
Hosting each group's complete expert pool on a single rank keeps dispatch and combine local, yielding equal per-rank input token counts and routing-independent communication under equal input partitions.
\gmoe's grouped expert placement complements MoonEP: when an expert group spans multiple ranks, our implementation applies MoonEP within each group, confining planning and expert migration to that group's ranks.

\section{Preliminaries}\label{sec:prelim}

Consider a single Mixture-of-Experts (MoE) layer applied after the multi-head attention block. Let $x_t \in \R^d$ denote the $d$-dimensional hidden representation of the $t$-th token entering the MoE module.

\paragraph{Standard MoE formulation.}
A conventional sparse MoE layer activates $K$ of $N$ routed experts for each token and computes its output as follows:
\begin{align}
\mathrm{MoE}(x_t) &= \sum_{i=1}^{N} g_i\,\mathrm{FFN}_i(x_t), \\
\mathcal{S}(x_t) &= \underset{j\in\{1,\ldots,N\}}{\operatorname{TopK}}\!\bigl(R(x_t)_j,\,K\bigr), \qquad g_i = \begin{cases} R(x_t)_i, & i\in\mathcal{S}(x_t), \\ 0, & \text{otherwise}.\end{cases}
\end{align}
Here, $\operatorname{TopK}(\cdot,K)$ returns the indices of the $K$ highest-scoring experts. The function $\mathrm{FFN}_i(\cdot)$ denotes the $i$-th expert network parameterized with an intermediate hidden dimension $d_{\mathrm{ffn}}$, and $R:\R^d\!\to\!\R^N$ represents the routing function (typically a softmax-normalized linear projection) that computes the token-to-expert affinity scores. To prevent routing collapse, the optimization objective is conventionally augmented with an auxiliary load-balancing loss~\citep{fedus2022switch}. For a batch or sequence of $T$ tokens, it is mathematically formulated as:
\begin{equation}
\mathcal{L}_{\mathrm{aux}} = \gamma \, N \sum_{i=1}^{N} f_i P_i, \quad \text{where} \quad f_i = \frac{1}{T}\sum_{t=1}^{T} \mathbb{I}\bigl\{i \in \mathcal{S}(x_t)\bigr\}, \quad P_i = \frac{1}{T}\sum_{t=1}^{T} R(x_t)_i,
\end{equation}
where $\mathcal{S}(x_t)$ denotes the set of $K$ selected expert indices for token $x_t$, $\mathbb{I}\{\cdot\}$ is the indicator function, $f_i$ is the dispatch fraction for expert $i$ over $T$ tokens, $P_i$ is its average routing probability, and $\gamma$ scales the auxiliary loss.

\paragraph{Standard fine-grained MoE.}
The fine-grained MoE architecture~\citep{dai2024deepseekmoe} introduces a structural degree of freedom by segmenting the intermediate hidden dimension of each expert by a factor $m$. This partition yields $mN$ finer-grained experts, each operating with a proportionally reduced intermediate dimension of $d_{\mathrm{ffn}}/m$. To strictly preserve the computational budget (FLOPs) of the standard MoE paradigm, the routing mechanism correspondingly activates $mK$ finer experts per token. Following DeepSeek-V3~\citep{deepseekv3} and LongCat-Flash~\citep{team2025longcat}, the routing formulation incorporates an additive load balancing bias term $b_j$~\citep{wang2024lossfree}:
\begin{align}
\mathrm{MoE}(x_t) &= \sum_{i=1}^{\Nexp} g_i\,\mathrm{FFN}_i(x_t), \\
\mathcal{S}(x_t) &= \underset{j\in\{1,\ldots,\Nexp\}}{\operatorname{TopK}}\!\bigl(R(x_t)_j+b_j,\,\kexp\bigr), \qquad g_i = \begin{cases} R(x_t)_i, & i\in\mathcal{S}(x_t), \\ 0, & \text{otherwise}. \end{cases}
\label{eq:finegrained-moe}
\end{align}
The quantities $\Nexp\!\equiv\!mN$ and $\kexp\!\equiv\!mK$ denote the total number of routed experts and the number selected per token in fine-grained MoE. We omit the $N_s$ always-on shared experts, which operate independently of routing.

\section{Grouped Mixture-of-Experts}\label{sec:method}

\subsection{Grouped Routing Architecture}\label{sec:method:arch}

Grouped Mixture-of-Experts (\gmoe) extends standard fine-grained MoE by introducing an orthogonal architectural dimension: the group count $\ngroup$. Each group retains the fine-grained MoE routing configuration of $\Nexp$ experts and $\kexp$ selections per token, with the fine-graining factor $m$ unchanged. Across the layer, this yields $\ngroup\Nexp$ routed experts and $\ngroup\kexp$ expert selections per token. Architecturally, as illustrated in Figure~\ref{fig:arch}, \gmoe deploys $\ngroup$ parallel instantiations of the fine-grained MoE. Each per-group expert operates on a partitioned subspace of dimension $\dsub = d/\ngroup$ while maintaining the original intermediate capacity of $d_{\mathrm{ffn}}/m$. Specifically, activating $\ngroup\kexp$ experts per token exactly matches the routed-expert FLOPs of the ungrouped counterpart, with additional computation from the input projection $W_{\mathrm{proj}}$ and merge projection $W_{\mathrm{merge}}$.

\paragraph{Input projection and split.}
Following the multi-head attention idiom, we mix features with the input projection $W_{\mathrm{proj}}$, apply RMSNorm to stabilize the subspace scale, and partition into $\ngroup$ contiguous chunks:
\begin{equation}
z_t = \mathrm{RMSNorm}\!\left(W_{\mathrm{proj}}\,x_t\right),\quad W_{\mathrm{proj}}\in\R^{d\times d},
\qquad
z_t = \bigl[z_{t,1};\,z_{t,2};\,\ldots;\,z_{t,\ngroup}\bigr],
\ \ z_{t,g}\in\R^{\dsub}.
\label{eq:gmoe-split}
\end{equation}

\paragraph{Independent per-group routing and expert computation.}
We apply forward scaling to each subspace input, yielding $\tilde{z}_{t,g} = \sqrt{\ngroup} z_{t,g}$.
Group $g$ owns its own softmax router $R_g:\R^{\dsub}\!\to\!\R^{\Nexp}$, its own fine-grained expert set $\{\mathrm{FFN}_{g,i}\}_{i=1}^{\Nexp}$ (each with intermediate dimension $d_{\mathrm{ffn}}/m$) and its own auxiliary-loss-free bias $\{b_{g,i}\}_{i=1}^{\Nexp}$. Its output is
\begin{align}
\mathcal{S}_g(x_t) &= \underset{j\in\{1,\ldots,\Nexp\}}{\operatorname{TopK}}\!\bigl(R_g(\tilde{z}_{t,g})_j+b_{g,j},\,\kexp\bigr), \qquad g_{g,i} = \begin{cases} R_g(\tilde{z}_{t,g})_i, & i\in\mathcal{S}_g(x_t),\\ 0, & \text{otherwise},\end{cases}\label{eq:gmoe-g}\\
y_{t,g} &= \sum_{i=1}^{\Nexp} g_{g,i}\,\mathrm{FFN}_{g,i}\!\left(\tilde{z}_{t,g}\right).
\label{eq:gmoe-group-out}
\end{align}

\paragraph{Merge projection and residual.}
The group outputs are concatenated back into a $d$-dimensional vector, mapped through the merge projection $W_{\mathrm{merge}}$, and combined with the shared experts:
\begin{equation}
\mathrm{GMoE}(x_t) = W_{\mathrm{merge}}\,\bigl[y_{t,1};\,\ldots;\,y_{t,\ngroup}\bigr]
              + \sum_{i=1}^{N_s} \mathrm{FFN}_i^{\mathrm{shared}}\!\left(x_t\right),
\qquad W_{\mathrm{merge}}\in\R^{d\times d}.
\label{eq:gmoe-merge}
\end{equation}

\subsection{Training-Aligned Parameterization}\label{sec:method:init}

Grouping reduces the fan-in of the router and expert gate/up projections from $d$ to $d/\ngroup$ and introduces an additional merge projection. We introduce a training-aligned parameterization that compensates for these changes while retaining the fine-grained baseline's training hyperparameters.

Using the $(a,b,c)$-parameterization framework underlying maximal update parameterization ($\mu$P)~\citep{yang2021tensor,everett2024scaling}, we express the scaling of the router and expert gate/up projections relative to the fine-grained baseline as
\begin{align}
h &= \ngroup^a W z_{t,g}, &
\sigma_{\mathrm{init}} &= \ngroup^b \sigma_{\mathrm{base}}, &
\eta &= \ngroup^c \eta_{\mathrm{base}}.
\label{eq:abc_def}
\end{align}
For each projection, $W$ denotes the trainable weight matrix and $h$ its pre-activation output. We denote the entrywise initialization standard deviation by $\sigma_{\mathrm{init}}$ and the learning rate by $\eta$. The exponents $a,b,c$ control forward, initialization, and learning-rate scaling, respectively. The parameterization combines three components:

\begin{itemize}[leftmargin=1.5em,itemsep=2pt]
    \item \textbf{Input-scale normalization.} The input projection $W_{\mathrm{proj}}$ changes the scale of the normalized hidden representation. We apply inner RMSNorm before splitting, as in~\eqref{eq:gmoe-split}, to normalize the projected representation. At initialization, each group input $z_{t,g}$ has approximately unit per-coordinate second moment, matching the input-scale reference of the fine-grained baseline.

    \item \textbf{Forward scaling for reduced fan-in.} At initialization, assume independent zero-mean weight entries that are independent of the normalized inputs. The pre-activation RMS relative to the baseline is
    \begin{equation}
    \frac{\mathrm{RMS}(h(0))}{\mathrm{RMS}(h_{\mathrm{base}}(0))}=\ngroup^{a+b-1/2}.
    \label{eq:init_scale_ratio}
    \end{equation}
    Without scaling, retaining the baseline initialization reduces this scale by $G^{-1/2}$.
Restoring the step-zero scale requires $a + b = 1/2$, which can be satisfied either by initialization-only scaling $(0, 1/2, 0)$ or by forward scaling $(1/2, 0, 0)$.
\gmoe adopts forward scaling, multiplying each group input by $\sqrt{G}$ while retaining the baseline initialization and learning rate, because it also compensates for parameter-induced updates as analyzed below.
The expert down projection retains fan-in $d_{\mathrm{ffn}}/m$, so its initial output scale is preserved when its input scale is matched.

    \item \textbf{Variance-preserving merge initialization.} The concatenated group outputs $y_t=[y_{t,1};\ldots;y_{t,\ngroup}]$ are mapped back to the residual stream by $W_{\mathrm{merge}}\in\R^{d\times d}$. We initialize its entries independently with zero mean and variance $1/d$. With aligned per-coordinate second moments $\mathbb{E}[y_{t,j}^2]=\sigma_y^2$, independence of the merge weights and their inputs gives
    \begin{equation}
    \mathbb{E}\bigl[(W_{\mathrm{merge}}y_t)_k^2\bigr]
    =\sum_{j=1}^{d}\frac{1}{d}\,\sigma_y^2=\sigma_y^2.
    \end{equation}
    This preserves the initial output variance through the merge projection and maintains the scale of the routed-expert contribution to the residual stream.
\end{itemize}

\paragraph{Update scaling under a shared learning rate.}
The forward multiplier also enters output changes induced by weight updates. Writing $z=z_{t,g}$, a single-step change decomposes as
\begin{equation}
\Delta h=\ngroup^a\bigl(\Delta W z+W\Delta z+\Delta W\Delta z\bigr).
\label{eq:full_delta_h}
\end{equation}
Here, $\ngroup^a W\Delta z$ propagates upstream representation changes through the current weights without reflecting updates to $W$ itself. By contrast, $\Delta h_W\triangleq\ngroup^a\Delta Wz$ isolates the direct effect of updating $W$ on the current input, and the cross term $\ngroup^a\Delta W\Delta z$ carries the same $\ngroup^a\Delta W$ factor. We therefore examine $\Delta h_W$ to compare the direct effects of weight updates under forward scaling and initialization scaling.

Under Adam with negligible stabilizer $\epsilon$, gradient normalization gives an entrywise update scale $\mathrm{RMS}(\Delta W)\asymp\eta=\ngroup^c\eta_{\mathrm{base}}$. Let $\alpha\in[1/2,1]$ describe how the contraction $\Delta Wz$ scales with fan-in at matched entrywise scales, ranging from uncorrelated accumulation to coherent alignment~\citep{everett2024scaling}. The baseline-relative update scale is
\begin{equation}
\frac{\mathrm{RMS}(\Delta h_W)}{\mathrm{RMS}(\Delta h_{W,\mathrm{base}})}\asymp\ngroup^{a+c-\alpha}.
\label{eq:update_ratio}
\end{equation}
For \gmoe, $(a,b,c)=(1/2,0,0)$ gives $\Delta h_W=\sqrt{\ngroup}\,\Delta Wz$ and a relative scale of $\ngroup^{1/2-\alpha}$. This ratio ranges from $1$ to $\ngroup^{-1/2}$ as $\alpha$ increases from $1/2$ to $1$. Table~\ref{tab:abc_scaling} compares parameterizations at the same learning rate. Initialization-only scaling $(0,1/2,0)$ matches the step-zero pre-activations but leaves the direct update ratio at $\ngroup^{-\alpha}$, illustrating the additional role of forward scaling in parameter-induced updates. Furthermore, measurements along the initialization-only trajectory show that update alignment shifts toward the square-root accumulation regime during training (Appendix~\ref{app:update-alignment}), supporting our choice of $\sqrt{\ngroup}$ rather than $\ngroup$ for forward scaling.

\begin{table}[!htbp]
\centering
\caption{Scaling factors and resulting initialization and update scales relative to the fine-grained baseline. Update ranges correspond to $\alpha\in[1/2,1]$.}
\label{tab:abc_scaling}
\begin{tabular}{lccc|cl@{}c@{}l}
\toprule
Parameterization & \multicolumn{3}{c|}{Scaling factors} & \multicolumn{4}{c}{Resulting scales} \\
\cmidrule(lr){2-4}\cmidrule(lr){5-8}
 & Forward $\ngroup^a$ & Init. std. $\ngroup^b$ & LR $\ngroup^c$ & Initial $h$ & \multicolumn{3}{c}{Update $\Delta h_W$} \\
\midrule
Unscaled (w/o $\sqrt{\ngroup}$) & $1$ & $1$ & $1$ & $\ngroup^{-1/2}$ & $\ngroup^{-\alpha}$ & $\;\in\;$ & $[\ngroup^{-1},\ngroup^{-1/2}]$ \\
Init-only scaling & $1$ & $\sqrt{\ngroup}$ & $1$ & $1$ & $\ngroup^{-\alpha}$ & $\;\in\;$ & $[\ngroup^{-1},\ngroup^{-1/2}]$ \\
\textbf{\gmoe (forward $\sqrt{\ngroup}$)} & $\boldsymbol{\sqrt{\ngroup}}$ & $\mathbf{1}$ & $\mathbf{1}$ & $\mathbf{1}$ & $\boldsymbol{\ngroup^{1/2-\alpha}}$ & $\;\in\;$ & $\boldsymbol{[\ngroup^{-1/2},1]}$ \\
\bottomrule
\end{tabular}
\end{table}

\subsection{Load-Balancing Calibration under Grouping}\label{sec:method:lb}

The architectural partitioning into $\ngroup$ subspaces inherently alters the relative scale of the gradient signals reaching the per-group routers. In this section, we analyze the backward pass dynamics at initialization to derive the recalibration factor required for the auxiliary load-balancing coefficient.

\paragraph{LM gradient attenuation.}
The language modeling (LM) objective reaches the per-group router through the backward pass of the merge projection. Let $\nabla_{\!\mathrm{GMoE}(x_t)} \mathcal{L}_{\mathrm{LM}}$ be the upstream gradient, which preserves its per-coordinate variance $\sigma_g^2$ at initialization due to the variance-preserving initialization of $W_{\mathrm{merge}}$ (Section~\ref{sec:method:init}). The gradient with respect to the pre-merged group output $y_{t,g} \in \R^{\dsub}$ is simply the corresponding chunk of $W_{\mathrm{merge}}^\top\nabla_{\!\mathrm{GMoE}(x_t)} \mathcal{L}_{\mathrm{LM}}$.

For a fixed token $t$, write $p_i^{(g)}=R_g(\tilde{z}_{t,g})_i$ and $e_i^{(g)}=\mathrm{FFN}_{g,i}(\tilde{z}_{t,g})$, suppressing the token index. The LM gradient applied to the routing probability $p_{i}^{(g)}$ is the inner product between this local gradient flow and the expert's forward activation $e_i^{(g)}\!\in\!\R^{\dsub}$:
\begin{equation}
\frac{\partial \mathcal{L}_{\mathrm{LM}}}{\partial p_{i}^{(g)}} = \sum_{k=1}^{\dsub} \bigl(\nabla_{\!y_{t,g}} \mathcal{L}_{\mathrm{LM}}\bigr)_k \cdot e^{(g)}_{i,k}.
\end{equation}
Under the standard analytical assumption that initialization yields zero-mean, uncorrelated forward activations and backward gradients, the variance of this inner product scales linearly with the subspace dimensionality $\dsub$. Consequently, the expected standard deviation of the LM gradient flowing into the router shrinks by a factor of $1/\sqrt{\ngroup}$ compared to the standard $d$-dimensional baseline:
\begin{equation}
\frac{\mathbb{E} \left\| \nabla_p \mathcal{L}_{\mathrm{LM}}^{(\mathrm{GMoE})} \right\|}{\mathbb{E} \left\| \nabla_p \mathcal{L}_{\mathrm{LM}}^{(\mathrm{base})} \right\|} = \frac{\sqrt{\dsub}}{\sqrt{d}} = \frac{1}{\sqrt{\ngroup}}.
\end{equation}

\paragraph{LB gradient invariance.}
We sum the auxiliary load-balancing losses across groups, using the same coefficient for each group. The gradient derived from the classical auxiliary load-balancing loss, $\mathcal{L}_{\mathrm{aux}} = \gamma \Nexp \sum_{g=1}^{\ngroup}\sum_{i=1}^{\Nexp} f_i^{(g)} P_i^{(g)}$, acts directly on the routing distribution. Because the $\sqrt{\ngroup}$ input scaling described in Section~\ref{sec:method:init} restores the pre-softmax logits to the expected statistical range of the ungrouped baseline at initialization, both the discrete routing frequencies $f_i^{(g)}$ and the continuous probabilities $P_i^{(g)}$ maintain their initial distributions. Thus, the expected magnitude of the auxiliary gradient remains independent of the spatial group count $\ngroup$:
\begin{equation}
\frac{\partial \mathcal{L}_{\mathrm{aux}}}{\partial p_i^{(g)}} \propto \gamma.
\end{equation}

\paragraph{Coefficient recalibration.}
Because the LM signal attenuates by $1/\sqrt{\ngroup}$ at initialization while the auxiliary signal remains static, the load-balancing penalty becomes implicitly $\sqrt{\ngroup}$ times stronger relative to the primary learning objective in \gmoe. Recent empirical studies on large-scale MoE training highlight that router stability fundamentally depends on the gradient norm ratio between the auxiliary load-balancing loss and the language modeling loss~\citep{team2025longcat}. If left uncalibrated, the relatively amplified load-balancing gradient would overwhelm the LM signal, forcing the router to distribute tokens uniformly regardless of the context and severely degrading conditional computation capabilities. 
To restore the optimization dynamics and preserve the baseline initial gradient norm ratio,  the auxiliary coefficient must be explicitly scaled down by this identical factor:
\begin{equation}
\gamma_{\mathrm{GMoE}} = \frac{\gamma_{\mathrm{base}}}{\sqrt{\ngroup}}.
\end{equation}

\paragraph{Compatibility with auxiliary-loss-free balancing.}
Auxiliary-loss-free balancing~\citep{wang2024lossfree} updates
the per-expert routing biases $b_{g,i}$ within each group using
observed expert loads. These biases
modify the scores used for top-$\kexp$ selection in
Eq.~\eqref{eq:gmoe-g} without introducing auxiliary balancing
gradients, so the gradient-based coefficient recalibration
above does not apply to their updates.
The training-aligned parameterization in
Section~\ref{sec:method:init} restores the initial router-logit
scale, while each group retains the baseline expert count
$\Nexp$ and selection budget $\kexp$, providing a consistent
scale reference for the additive routing biases.
With this alignment, we reuse the baseline bias update rate
without rescaling.

\section{Experiments}\label{sec:exp}

\subsection{Design Choices in \gmoe}\label{sec:analysis:design}
We validate the alignment mechanisms and projection design of \gmoe at a scale of $500$M activated and $10.5$B total parameters ($\ngroup{=}8$).

\subsubsection{Training Alignment and Load-Balancing Calibration}\label{sec:analysis:compatibility}
We first empirically validate the two alignment mechanisms introduced in Section~\ref{sec:method}, which share the primary goal of aligning \gmoe with the standard fine-grained MoE to enable seamless, zero-cost hyperparameter transfer. Specifically, the training-aligned parameterization (Section~\ref{sec:method:init}) addresses both initialization and update scales, while the auxiliary-loss recalibration (Section~\ref{sec:method:lb}) aligns the load-balancing strength with the grouped routing dynamics.

\begin{figure}[!htb]
\centering
\begin{subfigure}{0.49\textwidth}
\centering
\includegraphics[width=\linewidth]{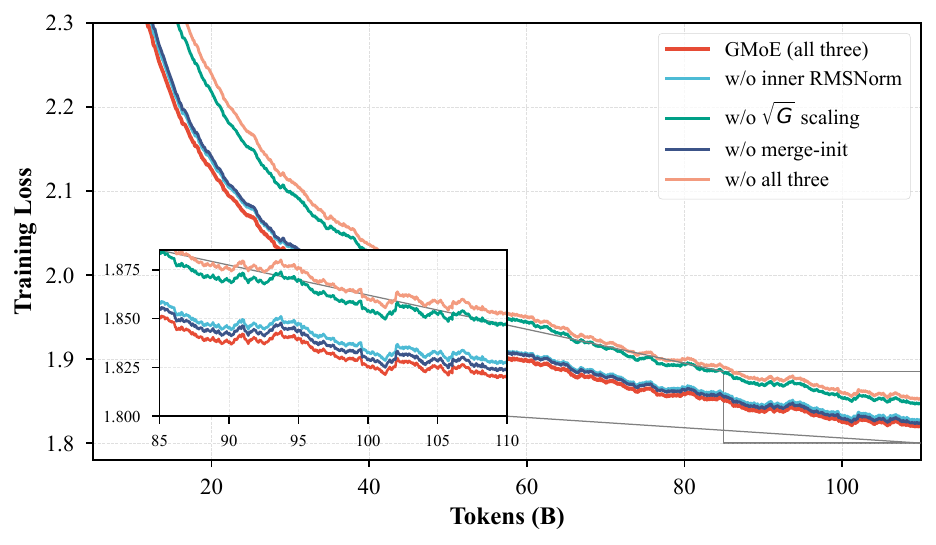}
\caption{Training-alignment ablation.}
\label{fig:align-ablation}
\end{subfigure}
\hfill
\begin{subfigure}{0.49\textwidth}
\centering
\includegraphics[width=\linewidth]{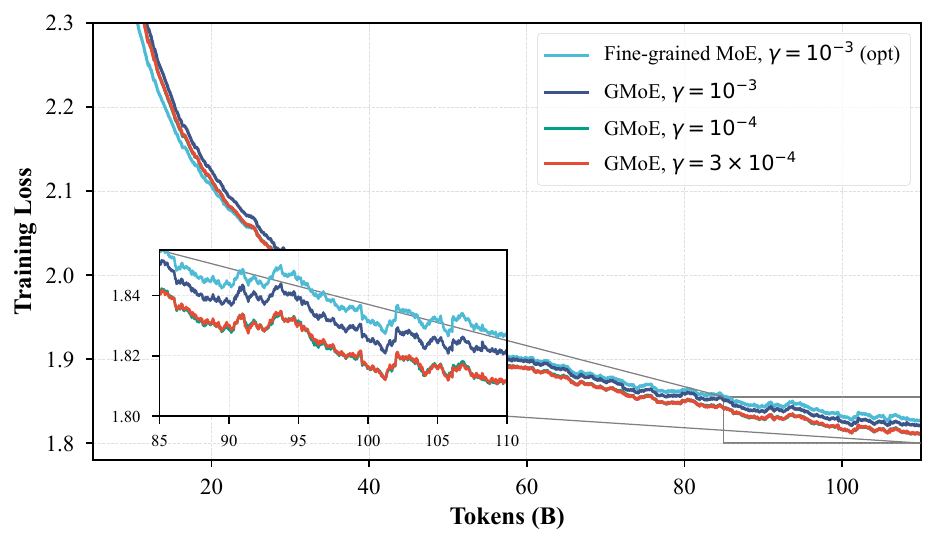}
\caption{Auxiliary-loss recalibration ablation.}
\label{fig:lb-calibration}
\end{subfigure}
\caption{Validating the design choices on \gmoe with $\ngroup{=}8$ ($500$M activated, $10.5$B total parameters). \textbf{(a)} Training-alignment ablation: the full alignment stays below every ablated variant, and removing all three choices yields the highest training loss. \textbf{(b)} Auxiliary-loss recalibration ablation: $\gamma{=}3\!\times\!10^{-4}$ reduces LM loss relative to directly reusing the baseline coefficient $10^{-3}$ and achieves a comparable loss to $10^{-4}$. All curves are EMA-smoothed.}
\label{fig:design-choices}
\end{figure}

\paragraph{Training-alignment ablation.}
We remove the three components of the training-aligned parameterization individually and jointly. Table~\ref{tab:align-ablation} reports training loss and validation loss at a matched $110$B-token budget, and Figure~\ref{fig:design-choices}\subref{fig:align-ablation} shows the corresponding training-loss trajectories. Removing any single component degrades both metrics, and disabling all three is the worst. Among the individual removals, the $\sqrt{\ngroup}$ forward scaling matters most. The full alignment achieves the best training and validation losses. Replacing $\sqrt{\ngroup}$ forward scaling with initialization-only scaling increases both training and validation loss despite matching the initial activation scale. The resulting losses nearly match those of the variant without $\sqrt{\ngroup}$ scaling in Table~\ref{tab:align-ablation} (Appendix~\ref{app:update-alignment}).

\begin{table}[!htb]
\centering
\small
\setlength{\tabcolsep}{8pt}
\renewcommand{\arraystretch}{1.08}
\caption{Ablation of the training-alignment design choices in \gmoe.}
\label{tab:align-ablation}
\begin{tabular}{lcc}
\toprule
Variant & Training loss $\downarrow$ & Validation loss $\downarrow$ \\
\midrule
\gmoe & \textbf{1.802} & \textbf{2.940} \\
\quad w/o inner RMSNorm & 1.809 & 2.952 \\
\quad w/o $\sqrt{\ngroup}$ scaling & 1.827 & 2.972 \\
\quad w/o merge initialization & 1.806 & 2.944 \\
\quad w/o all three & 1.834 & 2.978 \\
\bottomrule
\end{tabular}
\end{table}

\begin{figure}[!htb]
\centering
\includegraphics[width=\linewidth]{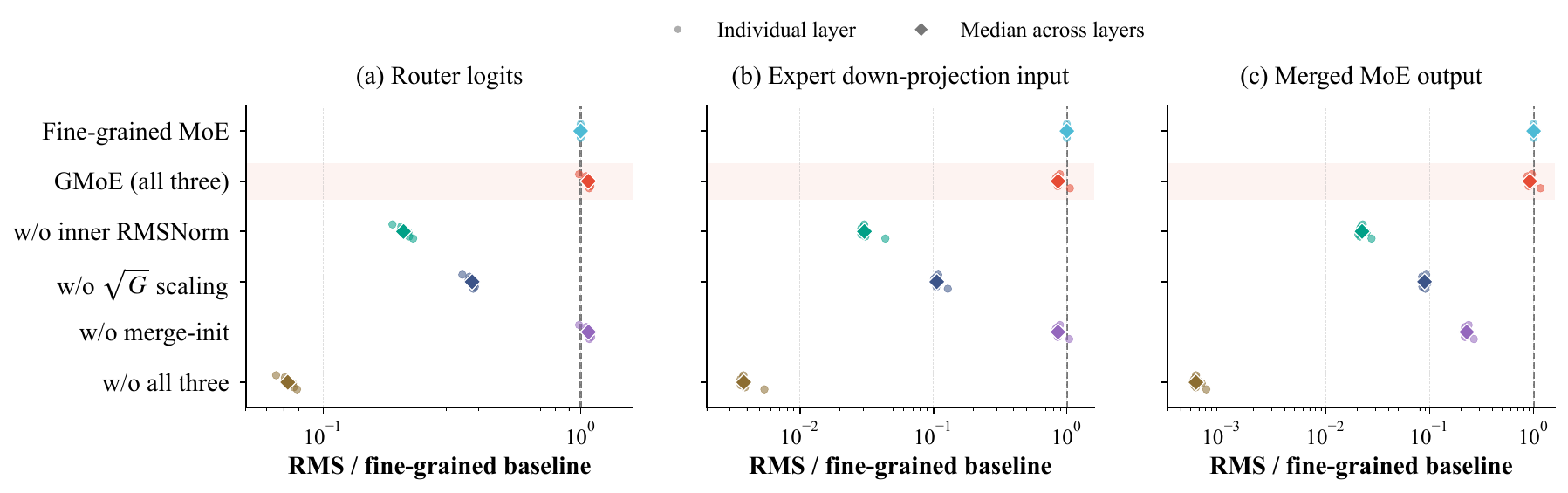}
\caption{Coordinate check of activation scales under training-alignment ablations ($\ngroup=8$) at training step 20. RMS statistics are normalized by the fine-grained baseline at the same layer. Circles show individual layers, and diamonds show the median across eight layers. Group-level statistics are averaged across groups before normalization. MoE outputs are measured before adding the shared-expert output and the residual.}
\label{fig:alignment-activation-scales}
\end{figure}

Figure~\ref{fig:alignment-activation-scales} compares activation scales across the alignment ablations early in training. With all three components, the RMS of router logits, expert down-projection inputs, and merged MoE outputs remains close to the fine-grained baseline across all eight layers. Removing inner RMSNorm suppresses all three scales. Removing the $\sqrt{\ngroup}$ forward scaling reduces router-logit RMS to $0.354$ of the complete configuration, consistent with $1/\sqrt{8}$. 
Removing merge initialization largely preserves the first two scales while substantially reducing the merged-output RMS.
These observations illustrate how the three components jointly align router logits, expert activations, and merged outputs with the fine-grained baseline.

\begin{figure}[htbp]
\centering
\includegraphics[width=0.9\linewidth]{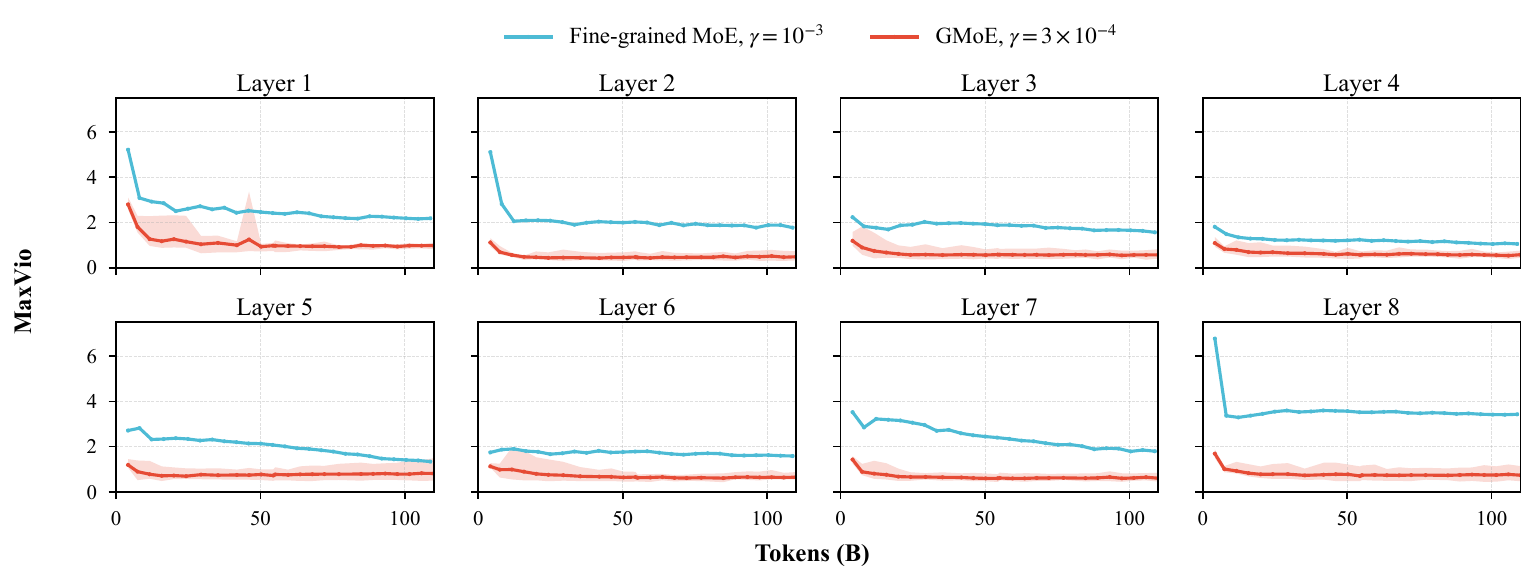}
\caption{Expert-level load balance after auxiliary-loss recalibration. For \gmoe, the line shows the mean of group-wise MaxVio and the shading spans the minimum and maximum across groups.}
\label{fig:expert-load-balance}
\end{figure}

\paragraph{Auxiliary-loss recalibration.}
Section~\ref{sec:method:lb} derives the recalibration $\gamma_{\mathrm{GMoE}}=\gamma_{\mathrm{base}}/\sqrt{\ngroup}$. For $\ngroup{=}8$ and the optimal baseline coefficient $\gamma_{\mathrm{base}}=10^{-3}$, we use $\gamma=3\!\times\!10^{-4}$. To validate this calibration, we compare it with direct transfer of the baseline coefficient and a further reduction to $10^{-4}$, testing whether a weaker auxiliary penalty provides additional benefits. 
As shown in Figure~\ref{fig:design-choices}\subref{fig:lb-calibration}, the calibrated coefficient achieves lower LM loss than direct transfer and matches $10^{-4}$ while maintaining stronger load-balancing regularization.
We also compare expert-level load balance between the baseline and \gmoe after auxiliary-loss recalibration, using maximal violation (MaxVio)~\citep{wang2024lossfree}. Figure~\ref{fig:expert-load-balance} shows lower group-averaged MaxVio for \gmoe than for the baseline across all eight layers. We hypothesize that separate routing subspaces reduce competition among heterogeneous token features, making expert utilization easier to balance even with a smaller auxiliary-loss coefficient.

\subsubsection{Subspace Projections}\label{sec:analysis:projection}
Compared with fine-grained MoE, \gmoe introduces an input projection $W_{\mathrm{proj}}$ and a merge projection $W_{\mathrm{merge}}$, raising the question of whether its gains can be attributed to the additional parameters. We construct two parameter-matched fine-grained MoE variants: one adds the same input and merge projections, and the other widens the shared experts. \gmoe achieves lower training and validation losses than both parameter-matched variants (Table~\ref{tab:projection-design-ablation}).

\begin{samepage}
We further investigate whether the projection parameter count can be reduced by sharing the input projection, merge projection, or both across groups. Sharing the input projection gives all groups the same projected representation, while sharing the merge projection maps their expert outputs back to the residual space through a common projection. Each sharing variant increases training and validation loss, with merge-projection sharing causing a larger increase than input-projection sharing. Sharing both yields a similar result to sharing the merge projection alone. These results support retaining group-specific mappings both into the routing subspaces and back to the residual space.\par
\end{samepage}

All \gmoe variants in Table~\ref{tab:projection-design-ablation} use the recalibrated auxiliary-loss coefficient.

\begin{table}[!htbp]
\centering
\small
\setlength{\tabcolsep}{8pt}
\renewcommand{\arraystretch}{1.0}
\caption{Ablation of the projection design in \gmoe.}
\label{tab:projection-design-ablation}
\begin{tabular}{lcc}
\toprule
Variant & Training loss $\downarrow$ & Validation loss $\downarrow$ \\
\midrule
\gmoe & \textbf{1.794} & \textbf{2.930} \\
Fine-grained MoE & 1.820 & 2.958 \\
\midrule
\multicolumn{3}{l}{\emph{Parameter-Matched Baselines}} \\
Fine-grained MoE + projections & 1.829 & 2.972 \\
Fine-grained MoE + wider shared experts & 1.819 & 2.957 \\
\midrule
\multicolumn{3}{l}{\emph{Projection sharing within} \gmoe} \\
\gmoe + shared input projection & 1.799 & 2.940 \\
\gmoe + shared merge projection & 1.803 & 2.947 \\
\gmoe + shared input/merge projections & 1.804 & 2.947 \\
\bottomrule
\end{tabular}
\end{table}

\subsection{Effect of Grouping on Sparsity Scaling}\label{sec:exp:scaling}

Sparse MoE can increase total model capacity by expanding the routed-expert pool while keeping the number and size of activated experts fixed. This provides a sparsity-scaling dimension at a constant routed-expert compute budget. We quantify sparsity as the ratio of total to activated routed experts, $s=\Nexp/\kexp$. In \gmoe, each group retains $\Nexp$ experts and selects $\kexp$ per token, giving $\ngroup\Nexp$ total experts and $\ngroup\kexp$ selections across the layer. The sparsity ratio therefore remains $s=\Nexp/\kexp$, providing a common basis for comparing the scaling behavior of fine-grained MoE and \gmoe.

At the same sparsity ratio, \gmoe offers a larger space of expert combinations through independent routing across groups. By partitioning routing across representational subspaces, this grouped architecture allows routers to capture differentiated semantic perspectives and produce complementary expert updates. Motivated by this, we evaluate whether grouping improves sparsity scaling compared with fine-grained MoE at the same routed-expert compute budget.

\paragraph{Experimental setup.}
We build a controlled scaling suite following the LongCat-Flash~\citep{team2025longcat} recipe with MLA attention, SwiGLU experts, RMSNorm, RoPE, and ScMoE routing placement. To isolate the effect of sparsity scaling, we hold the activated parameter budget ($\sim$500M) and active expert count ($k'=6$) fixed and sweep the expert-count configuration $\Nexp \in \{128, 256, 512, 1024, 2048\}$, increasing the total parameter count from $3.07$B to $37.04$B. All \gmoe models in this sweep use $\ngroup=4$ (Table~\ref{tab:scaling-backbones}).

\begin{table}[!htbp]
\centering
\small
\setlength{\tabcolsep}{5pt}
\renewcommand{\arraystretch}{1.15}
\caption{Configuration of the sparsity-scaling suite. All models are trained with $\sim\!500$M activated parameters.}
\label{tab:scaling-backbones}
\begin{tabular}{ccccccccccl}
\toprule
$L^{*}$ & Hidden & Head & $\Nexp$ & $\kexp$ & $d_e^{\dagger}$ & Batch$^{\ddagger}$ & LR & Tokens & Scan Values \\
\midrule
$10$ & $1536$ & $16$ & varies & $6$ & $384$ & $512$ & $4\!\times\!10^{-4}$ & $250$B & \makecell[l]{$\Nexp\!\in\!\{128,256,512,1024,2048\}$,\\ Total Params $\in\!\{3.07,5.33,9.86,18.92,37.04\}$B} \\
\bottomrule
\end{tabular}
\\[2pt]
\begin{minipage}{\linewidth}
\footnotesize\raggedright
$^{*}$ $L$ denotes the number of shortcut layers.\quad
$^{\dagger}$ $d_e=d_{\mathrm{ffn}}/m$ denotes the expert intermediate dimension.\quad
$^{\ddagger}$ All models were trained with a context length of 8192.
\end{minipage}
\end{table}

\begin{figure}[!htbp]
\centering
\definecolor{scalingbaseline}{HTML}{4DBBD5}
\definecolor{scalinggmoe}{HTML}{E64B35}
\begin{tikzpicture}
\begin{axis}[
  name=scalingaxis,
  width=0.72\linewidth,height=6.5cm,
  xlabel={Expert count $\Nexp$ (per group for GMoE)},ylabel={Validation Loss},
  xtick pos=bottom,
  xmode=log,log basis x=2,
  xtick={21.333333333333,42.666666666667,85.333333333333,170.666666666667,341.333333333333},
  xticklabels={128,256,512,1024,2048},
  xmin=18.571745350317,xmax=392.089038505654,ymin=2.80,ymax=3.02,
  legend style={font=\small,at={(0.98,0.98)},anchor=north east,draw=none,fill=white},
  grid=both,grid style={gray!25,dashed,line width=0.4pt},
  tick label style={font=\small},
  label style={font=\small\bfseries},
  every axis plot/.append style={line width=1.1pt,mark size=2pt},
]
\addplot[scalingbaseline,domain=21.333333333333:341.333333333333,samples=100,dashed,forget plot]
  {3.151830346444843*pow(x,-0.017)};
\addplot[scalinggmoe,domain=21.333333333333:341.333333333333,samples=100,dashed,forget plot]
  {3.120919599049882*pow(x,-0.017)};
\addplot[scalingbaseline,mark=*,only marks] coordinates {
  (21.333333333333,2.9971) (42.666666666667,2.9566) (85.333333333333,2.9129) (170.666666666667,2.8841) (341.333333333333,2.8590)
};
\addlegendentry{Fine-grained MoE: $\mathcal{L}=3.152s^{-0.017}$}
\addplot[scalinggmoe,mark=square*,only marks] coordinates {
  (21.333333333333,2.9661) (42.666666666667,2.9279) (85.333333333333,2.8883) (170.666666666667,2.8563) (341.333333333333,2.8320)
};
\addlegendentry{GMoE: $\mathcal{L}=3.121s^{-0.017}$}
\pgfmathsetmacro{\advantageloss}{2.89370038759414}
\draw[->,>=Stealth,thick]
  (axis cs:85.333333333333,\advantageloss) -- (axis cs:152.369572207379,\advantageloss)
  node[midway,above=3pt,font=\small,fill=white,inner sep=1pt] {$1.79\times$};
\end{axis}
\begin{axis}[
  at={(scalingaxis.south west)},anchor=south west,
  width=0.72\linewidth,height=6.5cm,
  axis x line*=top,axis y line=none,
  xlabel={Sparsity $s$},
  xmode=log,log basis x=2,
  xtick={21.333333333333,42.666666666667,85.333333333333,170.666666666667,341.333333333333},
  xticklabels={21.3,42.7,85.3,170.7,341.3},
  xmin=18.571745350317,xmax=392.089038505654,ymin=2.80,ymax=3.02,
  tick label style={font=\footnotesize,text=black!65},
  label style={font=\footnotesize,text=black!65},
  axis line style={draw=black!65},
  tick style={draw=black!65},
]
\end{axis}
\end{tikzpicture}
\caption{Sparsity scaling curves for \gmoe. \gmoe consistently outperforms fine-grained MoE across all sparsity levels. The fitted curves indicate that fine-grained MoE requires approximately $1.79\times$ the sparsity to achieve the same validation loss.}
\label{fig:scaling}
\end{figure}
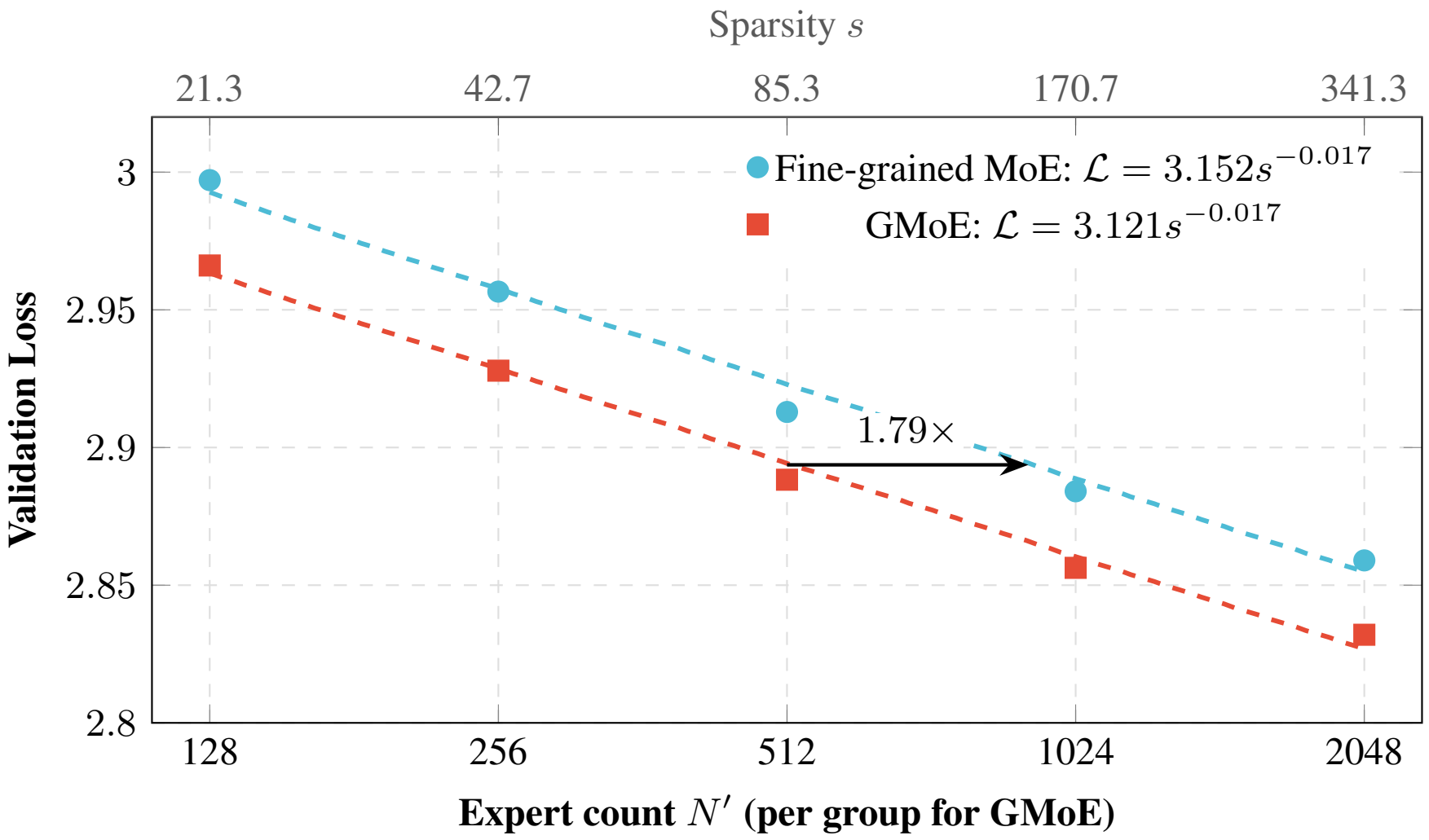

\paragraph{Grouping improves sparsity scaling.}
Figure~\ref{fig:scaling} plots validation loss across the five sparsity levels. \gmoe improves loss at every displayed scale, with absolute reductions of $0.025$ to $0.031$ nats across models with $3.07$B to $37.04$B total parameters. The two trajectories have similar slopes, while \gmoe consistently achieves lower loss.

The fitted curves are $\mathcal{L}_{\mathrm{base}}(s)=3.152s^{-0.017}$ and $\mathcal{L}_{\mathrm{GMoE}}(s)=3.121s^{-0.017}$ (Figure~\ref{fig:scaling}). Thus, fine-grained MoE requires approximately $1.79\times$ the sparsity of \gmoe to achieve the same validation loss:
\begin{equation}
\frac{s_{\mathrm{base}}}{s_{\mathrm{GMoE}}}
=\left(\frac{3.152}{3.121}\right)^{1/0.017}
\approx 1.79.
\label{eq:sparsity-ratio}
\end{equation}

\subsection{Large-scale Pre-Training}\label{sec:exp:main}

\paragraph{Training setup.}
The fine-grained baseline follows the LongCat-Flash-Lite backbone~\citep{oe2025} with 14 shortcut layers and MLA attention. Each MoE module contains 256 FFN experts and 128 zero-experts, with 12 experts selected per token. We construct \gmoe-69B by replacing these modules with the grouped architecture in Section~\ref{sec:method}, using $\ngroup=4$. The model has 69B total parameters and 3.8B activated parameters per token. The $N$-gram embedding contributes 31.4B parameters, accounting for $\sim\!46\%$ of the total parameter budget. We pre-train \gmoe-69B and the fine-grained baseline on a 5.0T-token corpus at a sequence length of 8192, using identical optimizer configurations, learning rate schedules, batch-size ramp-up, and data mixtures (Appendix~\ref{app:training_config}).

\begin{figure}[!htbp]
\centering
\includegraphics[width=0.7\linewidth]{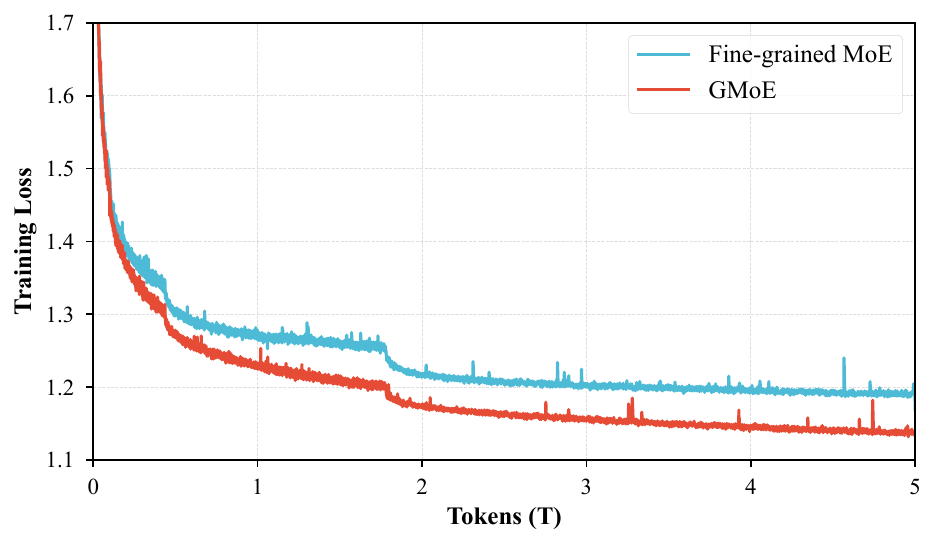}
\caption{LM loss vs.\ consumed tokens for \gmoe-69B and the fine-grained baseline. Curves are EMA-smoothed. Both runs share the same optimizer, data mixture, and training schedule. The loss drop coincides with a scheduled batch-size increase.}
\label{fig:lm-loss-3b}
\end{figure}

\paragraph{Base model evaluation.}
As illustrated in Figure~\ref{fig:lm-loss-3b}, \gmoe-69B maintains a strictly lower pre-training loss than the fine-grained baseline across the entire training trajectory. Beyond the $1$T token mark, \gmoe maintains a persistent loss advantage over the fine-grained baseline.

To assess downstream performance we evaluate both checkpoints on benchmarks spanning three areas:
\begin{itemize}
  \item \textbf{General:} MMLU~\citep{hendrycks2021mmlu}, MMLU-Pro~\citep{wang2024mmlupro}, GPQA-Diamond~\citep{rein2024gpqa}, BBH~\citep{suzgun2023bbh}, HellaSwag~\citep{zellers2019hellaswag}.
  \item \textbf{Math \& Code:} GSM8K~\citep{cobbe2021gsm8k}, MGSM~\citep{shi2022language}, MATH~\citep{hendrycks2021measuring}, CMATH~\citep{wei2023cmath}, HumanEval+~\citep{liu2023your}, MBPP+~\citep{liu2023your}.
  \item \textbf{Chinese:} C-Eval~\citep{huang2023ceval}, CMMLU~\citep{li2023cmmlu}.
\end{itemize}
We use the same evaluation settings across models and pre-training stages, with shot counts reported alongside the results. MMLU-Pro, GSM8K, MGSM, MATH, and CMATH use chain-of-thought prompting.

Table~\ref{tab:upscale-base} reports the comparison at $5$T pre-training tokens. \gmoe-69B improves over the fine-grained baseline on all thirteen benchmarks, with the largest gains concentrated in mathematical reasoning (GSM8K $+5.46$, MGSM $+4.50$), code generation (HumanEval+ $+4.27$), graduate-level reasoning (GPQA-Diamond $+5.62$), and commonsense reasoning (HellaSwag $+5.50$), together with consistent gains on English knowledge (MMLU $+3.46$, MMLU-Pro $+4.43$, BBH $+3.65$) and Chinese knowledge (C-Eval $+3.35$, CMMLU $+2.83$).

\begin{table}[!htbp]
\centering
\small
\caption{Base-model comparison between \gmoe-69B and the fine-grained baseline at $5$T pre-training tokens. Best per-row results are \textbf{bolded}.}
\label{tab:upscale-base}
\begin{tabular}{llc|cc}
\toprule
 & Benchmark          & \# Shots & Fine-grained-69B@5T & \gmoe-69B@5T \\
\midrule
\multirow{5}{*}{\emph{General}}
    & MMLU             & 5 & $70.90$          & $\mathbf{74.36}$ \\
    & MMLU-Pro         & 5 & $43.21$          & $\mathbf{47.64}$ \\
    & GPQA-Diamond     & 5 & $30.34$          & $\mathbf{35.96}$ \\
    & BBH              & 5 & $49.10$          & $\mathbf{52.75}$ \\
    & HellaSwag        & 5 & $62.50$          & $\mathbf{68.00}$ \\
\midrule
\multirow{6}{*}{\emph{Math \& Code}}
    & GSM8K            & 5 & $67.44$          & $\mathbf{72.90}$ \\
    & MGSM             & 8 & $58.50$          & $\mathbf{63.00}$ \\
    & MATH             & 5 & $29.22$          & $\mathbf{30.68}$ \\
    & CMATH            & 5 & $76.97$          & $\mathbf{77.31}$ \\
    & HumanEval+       & 3 & $55.49$          & $\mathbf{59.76}$ \\
    & MBPP+            & 3 & $57.41$          & $\mathbf{60.05}$ \\
\midrule
\multirow{2}{*}{\emph{Chinese}}
    & C-Eval           & 5 & $76.11$          & $\mathbf{79.46}$ \\
    & CMMLU            & 5 & $77.35$          & $\mathbf{80.18}$ \\
\bottomrule
\end{tabular}
\end{table}

\paragraph{Comparison with LatentMoE.}
Both LatentMoE~\citep{elango2026latentmoe} and \gmoe decouple the representation width processed by experts from the residual dimension $d$. LatentMoE achieves this through a shared latent space of width $\ell=d/\rho$, while retaining a single router on the original hidden representation. The reduced width lowers computation per expert, allowing the expert count and top-$k$ to increase by a factor of $\rho$. \gmoe instead partitions the projected representation into multiple subspaces, each with its own router and expert pool, enabling different subspaces to select their expert combinations independently.

We train a LatentMoE-69B variant using the same backbone, data mixture, and training schedule, and compare both models at $\sim\!1.4$T tokens. As shown in Table~\ref{tab:latentmoe-compare}, \gmoe-69B achieves superior performance on 10 out of 13 benchmarks, with notable improvements in commonsense reasoning (HellaSwag $+6.50$), code generation (HumanEval+ $+4.88$), Chinese knowledge (C-Eval $+2.95$), and mathematics (GSM8K $+2.77$).

\begin{table}[!htbp]
\centering
\small
\caption{Comparison between \gmoe-69B and LatentMoE-69B at $\sim\!1.4$T pre-training tokens, both after the same pre-training recipe. Best per-row results are \textbf{bolded}.}
\label{tab:latentmoe-compare}
\begin{tabular}{llc|cc}
\toprule
 & Benchmark          & \# Shots & LatentMoE-69B@1.4T & \gmoe-69B@1.4T \\
\midrule
\multirow{5}{*}{\emph{General}}
    & MMLU             & 5 & $68.82$          & $\mathbf{69.53}$ \\
    & MMLU-Pro         & 5 & $39.18$          & $\mathbf{40.04}$ \\
    & GPQA-Diamond     & 5 & $26.97$          & $\mathbf{27.53}$ \\
    & BBH              & 5 & $\mathbf{46.69}$ & $46.44$ \\
    & HellaSwag        & 5 & $50.00$          & $\mathbf{56.50}$ \\
\midrule
\multirow{6}{*}{\emph{Math \& Code}}
    & GSM8K            & 5 & $59.89$          & $\mathbf{62.66}$ \\
    & MGSM             & 8 & $\mathbf{53.80}$ & $53.30$ \\
    & MATH             & 5 & $\mathbf{22.48}$ & $20.64$ \\
    & CMATH            & 5 & $71.86$          & $\mathbf{73.54}$ \\
    & HumanEval+       & 3 & $42.68$          & $\mathbf{47.56}$ \\
    & MBPP+            & 3 & $56.35$          & $\mathbf{57.14}$ \\
\midrule
\multirow{2}{*}{\emph{Chinese}}
    & C-Eval           & 5 & $72.36$          & $\mathbf{75.31}$ \\
    & CMMLU            & 5 & $74.99$          & $\mathbf{76.03}$ \\
\bottomrule
\end{tabular}
\end{table}

\section{Analysis}\label{sec:analysis}

\subsection{Do Grouping Gains Come Solely from Finer Granularity?}\label{sec:analysis:fixed}

Fine-grained MoE architectures scale routing granularity and combinatorial expressivity by segmenting the FFN intermediate dimension while having each expert process the full hidden state via a single router. A foundational question arises regarding \gmoe: is hidden-space grouping merely another way to achieve finer expert granularity, or does it introduce a distinct structural advantage?

To isolate the structural effect of hidden-space grouping from FFN fine-graining, we conduct a controlled experiment under matched routed-expert capacity. For \gmoe, we use the $\Nexp=128$, $\ngroup=4$ configuration from Section~\ref{sec:exp:scaling}. In contrast, the fine-grained baseline retains the monolithic $1536$-dimensional space and employs a single router to select $24$ of $512$ experts, compressing the intermediate width of each expert to $96$. Aggregating over the entire layer, both designs instantiate exactly $512$ total experts, conditionally activate $24$ experts per token, and match both the parameter count and GEMM FLOPs of the routed experts.

\begin{figure}[!htbp]
\centering
\includegraphics[width=0.49\textwidth]{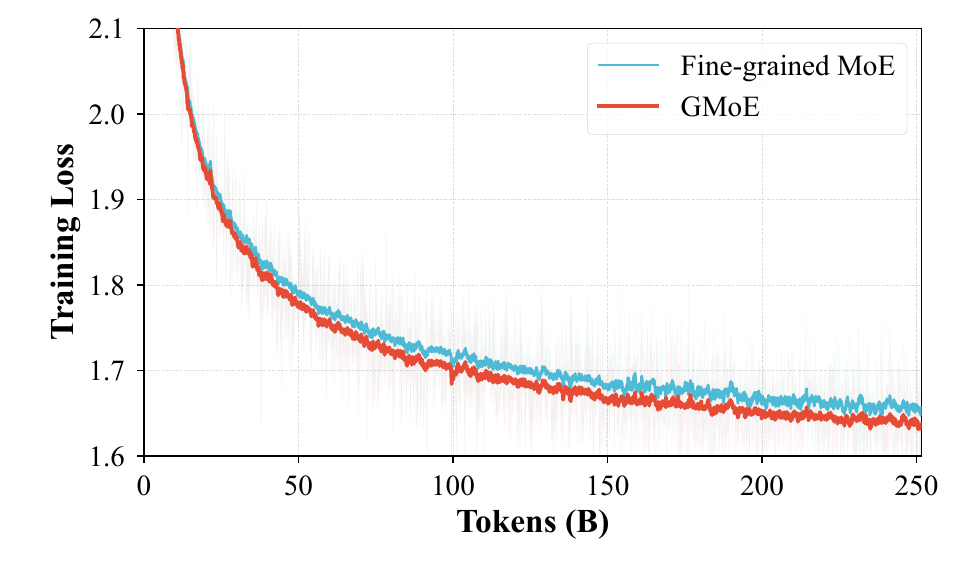}
\hfill
\includegraphics[width=0.49\textwidth]{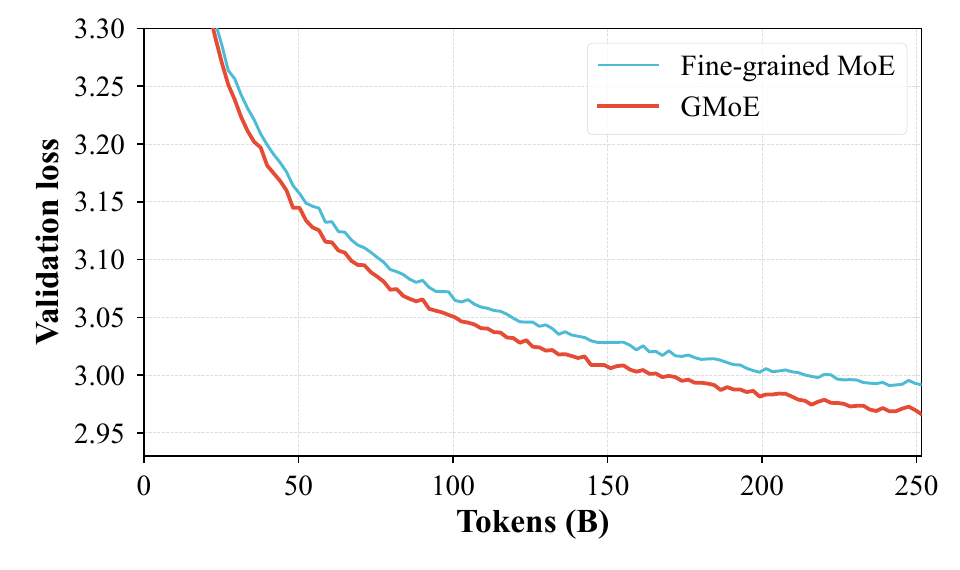}
\caption{Comparison of \gmoe and fine-grained MoE at matched routing granularity. Both models use $512$ routed experts and activate $24$ per token, with matched routed-expert parameters and GEMM FLOPs. \gmoe achieves lower training loss (left) and validation loss (right).}
\label{fig:fixed-expert}
\end{figure}

As illustrated in Figure~\ref{fig:fixed-expert}, \gmoe establishes a distinct advantage early in training and maintains a strictly lower training and validation loss. These results show that hidden-space grouping provides modeling benefits beyond further FFN fine-graining, even when the total and activated expert counts are matched.

\subsection{Where Does the Grouping Advantage Come From?}\label{sec:analysis:mechanism}

We examine the source of the grouping advantage using the four-group \gmoe-69B model introduced in Section~\ref{sec:exp:main}. We characterize the mechanism of \gmoe across three computational stages: determining whether $W_{\mathrm{proj}}$ yields overlapping or disjoint group representations, examining how independent routers make distinct routing decisions from group-specific projections of a shared token representation, and assessing whether the resulting expert updates are complementary or redundant.

\paragraph{Overlapping Group Representations.}
Although $W_{\mathrm{proj}}$ splits the projected representation into $\ngroup$ contiguous chunks, it does not isolate semantic information into strictly disjoint subspaces. Each group constructs its representation from the complete hidden state, producing overlapping views of the same token. We quantify this using linear centered kernel alignment (CKA)~\citep{kornblith2019similarity}, which compares the centered pairwise token similarity matrices induced by two representations (Appendix~\ref{app:group-analysis}). A value of $1$ indicates identical token geometry and $0$ indicates no shared linear structure. As summarized in Table~\ref{tab:group-cka-stages}, the group inputs exhibit high cross-group CKA ($0.816$), confirming that $W_{\mathrm{proj}}$ produces overlapping, group-specific views of the same token rather than disjoint feature slices.

\begin{table}[!htbp]
\centering
\small
\setlength{\tabcolsep}{9pt}
\caption{Cross-group geometry throughout the \gmoe computation. The reported values represent the mean pairwise linear CKA across all layers, evaluated on $50{,}000$ validation-set tokens.}
\label{tab:group-cka-stages}
\begin{tabular}{lcl}
\toprule
Stage & Cross-group CKA & Interpretation \\
\midrule
Group representations & $0.816$ & Shared token geometry \\
Routing weights        & $0.365$ & Differentiated routing \\
Group outputs          & $0.092$ & Distinct residual updates \\
\bottomrule
\end{tabular}
\end{table}

\paragraph{Group-Specific Routing.}
Despite their similar input geometries, the groups exhibit a sharp divergence in their token-routing behaviors. For routing CKA, each token is represented by its sparse top-$\kexp$ routing-weight vector, normalized to sum to one over the selected experts. The cross-group CKA of the routing weights drops significantly to $0.365$ (Table~\ref{tab:group-cka-stages}), revealing substantially different token-to-expert routing preferences. Table~\ref{tab:routing-examples} qualitatively illustrates these distinct semantic perspectives using seven anchor tokens. For the token ``government'', the four groups cluster neighbors based on broad societal contexts (``social'', ``military''), institutional structures (``congress'', ``council''), policy content (``law'', ``drug''), and political processes (``political'', ``vote''). These localized neighborhoods (defined formally in Appendix~\ref{app:routing_neighbors}) effectively expose complementary facets of a single underlying concept.

\begin{table}[!htbp]
\centering
\footnotesize
\setlength{\tabcolsep}{3.5pt}
\caption{Examples of routing views learned by different groups in \gmoe-69B.}
\label{tab:routing-examples}
\begin{adjustbox}{max width=\linewidth}
\begin{tabular}{lcccc}
\toprule
Anchor & Group 0 & Group 1 & Group 2 & Group 3 \\
\midrule
\emph{government}
& \makecell{\emph{social}, \emph{military},\\\emph{economic}}
& \makecell{\emph{congress}, \emph{council},\\\emph{committee}}
& \makecell{\emph{law}, \emph{drug},\\\emph{policy}}
& \makecell{\emph{political}, \emph{vote},\\\emph{campaign}} \\
\emph{language}
& \makecell{\emph{word}, \emph{name},\\\emph{term}}
& \makecell{\emph{text}, \emph{music},\\\emph{letters}}
& \makecell{\emph{culture}, \emph{history},\\\emph{religion}}
& \makecell{\emph{nature}, \emph{mental},\\\emph{ideas}} \\
\emph{number}
& \makecell{\emph{kinds}, \emph{types},\\\emph{amount}}
& \makecell{\emph{series}, \emph{parts},\\\emph{pieces}}
& \makecell{\emph{two}, \emph{three},\\\emph{thousands}}
& \makecell{\emph{rate}, \emph{average},\\\emph{value}} \\
\emph{written}
& \makecell{\emph{write}, \emph{published},\\\emph{quoted}}
& \makecell{\emph{taught}, \emph{spoke},\\\emph{learned}}
& \makecell{\emph{reading}, \emph{stories},\\\emph{translation}}
& \makecell{\emph{story}, \emph{speech},\\\emph{manuscript}} \\
\emph{new}
& \makecell{\emph{old}, \emph{previous},\\\emph{fresh}}
& \makecell{\emph{real}, \emph{best},\\\emph{perfect}}
& \makecell{\emph{build}, \emph{built},\\\emph{create}}
& \makecell{\emph{changing}, \emph{change},\\\emph{reform}} \\
\begin{CJK*}{UTF8}{gbsn}电力系统\end{CJK*}
& \makecell{\begin{CJK*}{UTF8}{gbsn}控制系统、监控系统、\end{CJK*}\\\begin{CJK*}{UTF8}{gbsn}物联网\end{CJK*}}
& \makecell{\begin{CJK*}{UTF8}{gbsn}电力设备、电气设备、\end{CJK*}\\\begin{CJK*}{UTF8}{gbsn}自动化设备\end{CJK*}}
& \makecell{\begin{CJK*}{UTF8}{gbsn}电气工程、土木工程、\end{CJK*}\\\begin{CJK*}{UTF8}{gbsn}建筑工程\end{CJK*}}
& \makecell{\begin{CJK*}{UTF8}{gbsn}电网、发电厂、\end{CJK*}\\\begin{CJK*}{UTF8}{gbsn}变电站\end{CJK*}} \\
\begin{CJK*}{UTF8}{gbsn}经济\end{CJK*}
& \makecell{\begin{CJK*}{UTF8}{gbsn}物质、大气、\end{CJK*}\\\begin{CJK*}{UTF8}{gbsn}地域\end{CJK*}}
& \makecell{\begin{CJK*}{UTF8}{gbsn}市场经济、历史、\end{CJK*}\\\begin{CJK*}{UTF8}{gbsn}人生\end{CJK*}}
& \makecell{\begin{CJK*}{UTF8}{gbsn}社会效益、管理水平、\end{CJK*}\\\begin{CJK*}{UTF8}{gbsn}经济效益\end{CJK*}}
& \makecell{\begin{CJK*}{UTF8}{gbsn}经济发展、资源、\end{CJK*}\\\begin{CJK*}{UTF8}{gbsn}土地资源\end{CJK*}} \\
\bottomrule
\end{tabular}
\end{adjustbox}
\end{table}

\paragraph{Cross-Group Complementarity.}
After projection back to the shared residual space, the group outputs exhibit a cross-group CKA of $0.092$ (Table~\ref{tab:group-cka-stages}), substantially lower than that of the group inputs. Thus, the groups transform overlapping input representations into residual updates with distinct token geometries. Consistent with this observation, sharing the merge projection causes a larger loss increase than sharing the input projection (Section~\ref{sec:analysis:projection}), supporting group-specific mappings back to the residual space.
Collectively, these observations characterize \gmoe through shared input geometry, differentiated routing, and complementary outputs.

\subsection{How Does Group Count Affect Quality and Efficiency?}\label{sec:analysis:gstar}

Having examined how grouping enables differentiated routing and produces complementary outputs, we next study how the group count $\ngroup$ affects validation quality and local execution efficiency. Building on the recipe from Section~\ref{sec:exp:scaling}, we sweep model widths $d\in\{768,1536,2304,3072\}$ and group counts $\ngroup\in\{1,2,4,8,16\}$. With the per-group expert count $\Nexp=128$, top-$\kexp=6$, and expert intermediate width $d_e=d/4$ held fixed within each width, varying $\ngroup$ changes the subspace width $d/\ngroup$ and the number of independent routers while preserving the total routed-expert parameter count and core FFN FLOPs per token.

\begin{figure}[htbp]
\centering
\definecolor{gstarblue}{RGB}{62,150,221}
\definecolor{gstarorange}{RGB}{230,126,55}
\definecolor{gstargreen}{RGB}{41,169,85}
\definecolor{gstarred}{RGB}{204,44,44}
\begin{tikzpicture}
\begin{axis}[
  width=0.91\linewidth,
  height=5.25cm,
  xmin=2.75,
  xmax=3.24,
  ymin=0,
  ymax=109,
  xtick={2.76,2.88,3.00,3.12,3.24},
  xticklabel style={/pgf/number format/fixed,/pgf/number format/fixed zerofill,/pgf/number format/precision=2},
  ytick={0,20,40,60,80,100},
  xlabel={Validation loss (lower is better)},
  ylabel={Relative FP8 utilization$^{*}$ (\% of $\ngroup=1$)},
  tick label style={font=\scriptsize},
  label style={font=\scriptsize},
  grid=both,
  major grid style={draw=gray!22,line width=0.35pt},
  minor grid style={draw=gray!10,line width=0.25pt},
  axis line style={draw=black!60},
  tick style={draw=black!60},
  clip=false,
  legend style={
    at={(0.5,1.03)},
    anchor=south,
    draw=none,
    fill=none,
    font=\scriptsize,
    legend columns=4,
    /tikz/every even column/.append style={column sep=0.8em},
  },
]

\addlegendimage{gstarred,line width=1.35pt}
\addlegendentry{$d=3072$}
\addlegendimage{gstargreen,line width=1.35pt}
\addlegendentry{$d=2304$}
\addlegendimage{gstarorange,line width=1.35pt}
\addlegendentry{$d=1536$}
\addlegendimage{gstarblue,line width=1.35pt}
\addlegendentry{$d=768$}

\addplot[gstarred,line width=1.35pt] coordinates {
  (2.8612,100.0000) (2.8345,92.1847) (2.8223,78.9601)
  (2.8203,59.8205) (2.8199,34.4085)
};
\addplot[only marks,mark=*,mark size=2.1pt,gstarred] coordinates {
  (2.8612,100.0000) (2.8345,92.1847) (2.8223,78.9601)
  (2.8203,59.8205) (2.8199,34.4085)
};

\addplot[gstargreen,line width=1.35pt] coordinates {
  (2.9193,100.0000) (2.8978,87.5167) (2.8827,67.8930) (2.8752,45.9891)
};
\addplot[gstargreen!65,densely dashed,line width=0.85pt] coordinates {
  (2.8752,45.9891) (2.8853,26.5668)
};
\addplot[only marks,mark=*,mark size=2.1pt,gstargreen] coordinates {
  (2.9193,100.0000) (2.8978,87.5167) (2.8827,67.8930)
  (2.8752,45.9891) (2.8853,26.5668)
};

\addplot[gstarorange,line width=1.35pt] coordinates {
  (3.0085,100.0000) (2.9864,83.7923) (2.9734,64.6400)
};
\addplot[gstarorange!65,densely dashed,line width=0.85pt] coordinates {
  (2.9734,64.6400) (2.9749,38.0740) (2.9861,22.1556)
};
\addplot[only marks,mark=*,mark size=2.1pt,gstarorange] coordinates {
  (3.0085,100.0000) (2.9864,83.7923) (2.9734,64.6400)
  (2.9749,38.0740) (2.9861,22.1556)
};

\addplot[gstarblue,line width=1.35pt] coordinates {
  (3.2196,100.0000) (3.1965,78.1046) (3.1936,46.6523)
};
\addplot[gstarblue!65,densely dashed,line width=0.85pt] coordinates {
  (3.1936,46.6523) (3.1975,27.3799) (3.2204,13.6895)
};
\addplot[only marks,mark=*,mark size=2.1pt,gstarblue] coordinates {
  (3.2196,100.0000) (3.1965,78.1046) (3.1936,46.6523)
  (3.1975,27.3799) (3.2204,13.6895)
};

\node[font=\scriptsize,anchor=east,xshift=-1.5pt] at (axis cs:2.8612,100.0000) {$\ngroup=1$};
\node[font=\scriptsize,anchor=east,xshift=-1.5pt] at (axis cs:2.8345,92.1847) {$\ngroup=2$};
\node[font=\scriptsize,anchor=east,xshift=-1.5pt] at (axis cs:2.8223,78.9601) {$\ngroup=4$};
\node[font=\scriptsize,anchor=east,xshift=-1.5pt] at (axis cs:2.8203,59.8205) {$\ngroup=8$};
\node[font=\scriptsize,anchor=east,xshift=-1.5pt] at (axis cs:2.8199,34.4085) {$\ngroup=16$};

\node[font=\scriptsize,anchor=west,xshift=1.5pt] at (axis cs:2.9193,100.0000) {$\ngroup=1$};
\node[font=\scriptsize,anchor=west,xshift=1.5pt] at (axis cs:2.8978,87.5167) {$\ngroup=2$};
\node[font=\scriptsize,anchor=west,xshift=1.5pt] at (axis cs:2.8827,67.8930) {$\ngroup=4$};
\node[font=\scriptsize,anchor=west,xshift=1.5pt] at (axis cs:2.8752,45.9891) {$\ngroup=8$};
\node[font=\scriptsize,anchor=west,xshift=1.5pt] at (axis cs:2.8853,26.5668) {$\ngroup=16$};

\node[font=\scriptsize,anchor=west,xshift=1.5pt] at (axis cs:3.0085,100.0000) {$\ngroup=1$};
\node[font=\scriptsize,anchor=west,xshift=1.5pt] at (axis cs:2.9864,83.7923) {$\ngroup=2$};
\node[font=\scriptsize,anchor=west,xshift=1.5pt] at (axis cs:2.9734,64.6400) {$\ngroup=4$};
\node[font=\scriptsize,anchor=west,xshift=1.5pt] at (axis cs:2.9749,38.0740) {$\ngroup=8$};
\node[font=\scriptsize,anchor=west,xshift=1.5pt] at (axis cs:2.9861,22.1556) {$\ngroup=16$};

\node[font=\scriptsize,anchor=east,xshift=-1.5pt] at (axis cs:3.2196,100.0000) {$\ngroup=1$};
\node[font=\scriptsize,anchor=east,xshift=-1.5pt] at (axis cs:3.1965,78.1046) {$\ngroup=2$};
\node[font=\scriptsize,anchor=east,xshift=-1.5pt] at (axis cs:3.1936,46.6523) {$\ngroup=4$};
\node[font=\scriptsize,anchor=east,xshift=-1.5pt] at (axis cs:3.1975,27.3799) {$\ngroup=8$};
\node[font=\scriptsize,anchor=east,xshift=-1.5pt] at (axis cs:3.2204,13.6895) {$\ngroup=16$};
\end{axis}
\end{tikzpicture}

{\captionsetup{font=scriptsize,justification=raggedright,singlelinecheck=false,width=0.91\linewidth,skip=3pt}
\caption*{Solid segments: within-width Pareto frontier. Dashed segments: dominated continuation.\par
$^{*}$\,Relative FP8 utilization is useful expert GEMM FLOPs divided by local MoE forward time, normalized to $\ngroup=1$ at each width. Timings use CUDA Graph replay and exclude input and merge projections and cross-rank communication.}}
\caption{Validation loss and relative FP8 utilization across group counts.}
\label{fig:gstar}
\end{figure}

Figure~\ref{fig:gstar} shows that the loss-minimizing group count shifts toward larger values as model width increases. At the same time, quality gains diminish beyond moderate group counts, and further grouping can increase validation loss in narrower models. Meanwhile, relative FP8 utilization declines as $\ngroup$ increases, despite unchanged routed-expert GEMM FLOPs. At $d=3072$, further grouping beyond $\ngroup=4$ yields only small quality gains at this training budget, while substantially reducing local execution efficiency under our benchmark conditions. These results inform our choice of $\ngroup=4$ for \gmoe-69B, balancing validation quality against local execution efficiency.

\section{System Optimization}\label{sec:system}
\gmoe provides a structural boundary for expert routing dynamism: every input token is processed by every group, while expert selection is confined to each group's independent pool.
For a fixed input batch, all groups receive the same number of subspace tokens and select the same number of experts per token, yielding equal and fixed numbers of token--expert assignments.
We exploit this structure by placing each group on a single device under expert parallelism (EP), balancing assignments across devices and eliminating routing-dependent dispatch/combine communication.
We further optimize FP8 expert computation through kernel fusion and intra-kernel pipelining, and exploit fixed device-level input shapes to simplify CUDA Graph capture and replay.

\subsection{Expert Parallelism with Group-Local Placement and Routing-Independent Communication}\label{sec:system:comm}

The $\ngroup \times \Nexp$ routed experts in \gmoe can be evenly distributed across $\mathrm{EP}$ ranks using expert parallelism.
Under group-distributed placement, with $\mathrm{EP}>\ngroup$ and $\ngroup$ dividing $\mathrm{EP}$, each group's expert pool spans $\mathrm{EP}/\ngroup$ ranks.
Routing decisions then affect both per-rank workloads and intra-group dispatch/combine all-to-allv communication.
We adopt group-local placement with $\mathrm{EP}=\ngroup$, placing each group's complete expert pool on a single rank.
This placement yields equal assignment counts across ranks and keeps dispatch and combine local, eliminating routing-dependent variable-size token communication between ranks.

The remaining token communication follows the fixed group layout. 
Before routing, an equal-partition all-to-all, \emph{sp2hp} (sequence-to-hidden-partition layout conversion), delivers subspace representations to their assigned ranks, converting each rank's input from $[S,d]$ to $[\ngroup S,d/\ngroup]$, where $S$ is the input token count per rank. After local expert computation, \emph{hp2sp} reverses this layout conversion.
Appendices~\ref{app:distributed_execution} and~\ref{app:comm_accounting} detail the (DP, CP) rank mapping and cross-rank communication volumes, respectively.

\begin{figure}[!htbp]
\centering
\includegraphics[width=\linewidth]{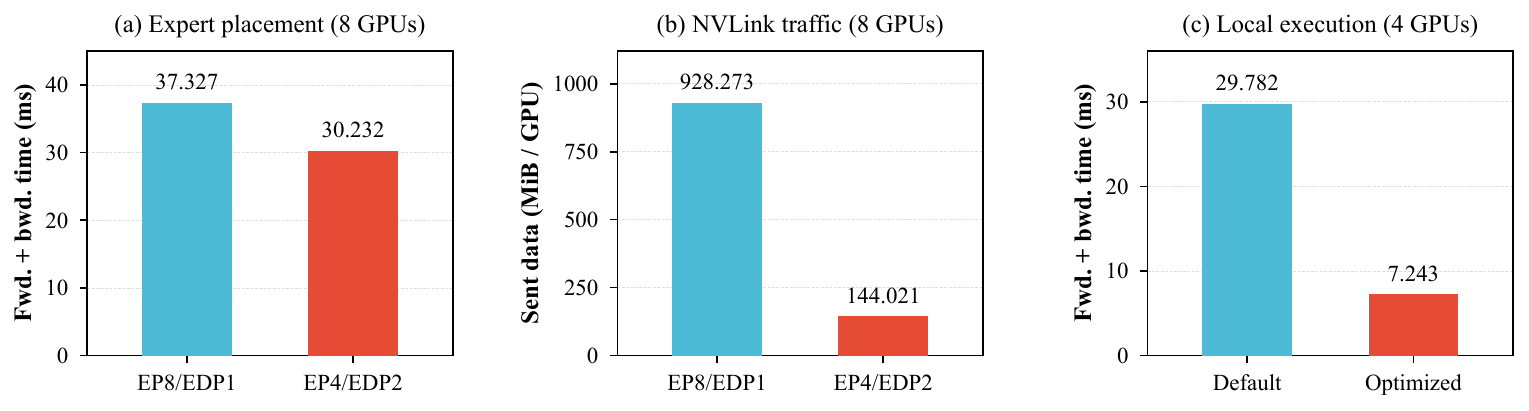}
\caption{
System performance of \gmoe. Group-local placement reduces module runtime and NVLink traffic on eight GPUs (a,b). With group-local placement, local execution optimization accelerates the module on four GPUs (c). Timings cover the full forward and backward passes of the module.}
\label{fig:system-performance}
\end{figure}

We evaluate group-local placement against group-distributed placement using the same trained four-group \gmoe with $d=3072$, fixed real inputs, and identical routing decisions on eight GPUs. With expert data parallelism (EDP), EP4/EDP2 places one complete group on each rank with two expert-data-parallel replicas per group, whereas EP8/EDP1 splits each group across two ranks. Figure~\ref{fig:system-performance}(a,b) shows that group-local placement reduces the module's combined forward and backward time from $37.327$ to $30.232$\,ms ($19.0\%$), while the separately measured NVLink send volume decreases from $928.273$ to $144.021$\,MiB per GPU ($84.5\%$). These savings come at the cost of storing twice as many expert parameters per rank, trading memory for local expert execution.

\subsection{Fused MoE Kernels with Static Buffers}\label{sec:system:local}
Compared with fine-grained MoE, \gmoe activates more experts with narrower input and output dimensions but unchanged intermediate width. This preserves the total expert GEMM FLOPs per token while increasing the intermediate activation volume and elementwise SwiGLU work, which can limit hardware utilization. We address these challenges by optimizing the FP8 expert computation through kernel fusion and intra-kernel pipelining. Kernel fusion reduces intermediate memory traffic and launch overhead, while multistage pipelining overlaps asynchronous data loads with computation to improve utilization of CUDA cores and Tensor Cores. Our implementation incorporates GEMM components adapted from DeepGEMM~\citep{deepgemm2025}.
Given fixed device-level input shapes, we use fixed-capacity buffers and GPU-resident routing metadata to handle local expert assignments. This simplifies CUDA Graph capture and replay of the forward and backward passes under expert parallelism.

To measure local execution gains under group-local placement, we compare the default and optimized implementations on four GPUs, using the same configuration. Figure~\ref{fig:system-performance}(c) shows a reduction in the module's combined forward and backward time from $29.782$ to $7.243$\,ms, a $4.11\times$ speedup. Separately measured forward time decreases from $10.206$ to $2.748$\,ms and backward time from $19.613$ to $4.515$\,ms.

\subsection{End-to-End Training Efficiency}\label{sec:system:training}

To assess the impact on complete training updates, we benchmark \gmoe-69B from random initialization on $8$, $16$, and $64$ GPUs, with a global batch size of $1024$ and a sequence length of $8192$. On eight GPUs, the default and optimized implementations are interleaved within the same GPU allocation. We report the median step time over steps $6$ to $12$ from two runs. 
Table~\ref{tab:system-training} shows that the optimized implementation reduces eight-GPU step time from $76.296$ to $42.716$ seconds, a $1.79\times$ speedup over the default \gmoe implementation, with throughput increasing from $109.95$k to $196.38$k tokens/s. At larger GPU counts, the optimized implementation reaches $383.23$k and $1380.48$k tokens/s on $16$ and $64$ GPUs, respectively, demonstrating efficient multi-GPU scaling.

\begin{table}[!htbp]
\centering
\small
\setlength{\tabcolsep}{9pt}
\caption{Training efficiency of \gmoe-69B.}
\label{tab:system-training}
\begin{tabular}{rlrr}
\toprule
GPUs & Implementation & Step time (s) & Throughput (k tokens/s) \\
\midrule
8 & Default & $76.296$ & $109.95$ \\
8 & Optimized & $\mathbf{42.716}$ & $\mathbf{196.38}$ \\
\midrule
16 & Optimized & $21.889$ & $383.23$ \\
64 & Optimized & $6.077$ & $1380.48$ \\
\bottomrule
\end{tabular}
\end{table}

\section{Conclusion}
In this work, we systematically investigate hidden-space grouping as an architectural and systems dimension for fine-grained MoE architectures through \gmoe. 
We address the fundamental training dynamics of grouped architectures by developing a training-aligned parameterization and load-balancing calibration, enabling zero-cost hyperparameter transfer from standard fine-grained baselines. 
Empirically, fine-grained MoE requires approximately $1.79\times$ higher sparsity to match \gmoe's validation loss, and \gmoe-69B outperforms its fine-grained baseline across all 13 downstream benchmarks after 5T tokens while surpassing LatentMoE on 10 of 13 benchmarks after 1.4T tokens.
At the systems level, aligning expert parallelism with the group layout $(\mathrm{EP}=\ngroup)$ keeps routing and expert computation local to each rank, eliminating intra-group dispatch and combine collectives while removing routing-dependent cross-rank token imbalances. 
Overall, these results establish \gmoe as an effective and scalable design principle, bridging model capacity and distributed execution efficiency for next-generation sparse architectures.

\bibliographystyle{plainnat}
\bibliography{references}

@inproceedings{hoffmann2022chinchilla,
  title={Training Compute-Optimal Large Language Models},
  author={Hoffmann, Jordan and Borgeaud, Sebastian and Mensch, Arthur and others},
  booktitle={Advances in Neural Information Processing Systems (NeurIPS)},
  year={2022}
}

@inproceedings{shazeer2017moe,
  title={Outrageously Large Neural Networks: The Sparsely-Gated Mixture-of-Experts Layer},
  author={Shazeer, Noam and Mirhoseini, Azalia and Maziarz, Krzysztof and Davis, Andy and Le, Quoc and Hinton, Geoffrey and Dean, Jeff},
  booktitle={International Conference on Learning Representations (ICLR)},
  year={2017}
}

@article{fedus2022switch,
  title={Switch Transformers: Scaling to Trillion Parameter Models with Simple and Efficient Sparsity},
  author={Fedus, William and Zoph, Barret and Shazeer, Noam},
  journal={Journal of Machine Learning Research (JMLR)},
  year={2022}
}

@inproceedings{lepikhin2021gshard,
  title={{GShard}: Scaling Giant Models with Conditional Computation and Automatic Sharding},
  author={Lepikhin, Dmitry and Lee, HyoukJoong and Xu, Yuanzhong and others},
  booktitle={International Conference on Learning Representations (ICLR)},
  year={2021}
}

@article{zoph2022stmoe,
  title={{ST-MoE}: Designing Stable and Transferable Sparse Expert Models},
  author={Zoph, Barret and Bello, Irwan and Kumar, Sameer and others},
  journal={arXiv preprint arXiv:2202.08906},
  year={2022}
}

@inproceedings{dai2024deepseekmoe,
  title={{DeepSeekMoE}: Towards Ultimate Expert Specialization in Mixture-of-Experts Language Models},
  author={Dai, Damai and Deng, Chengqi and Zhao, Chenggang and others},
  booktitle={Annual Meeting of the Association for Computational Linguistics (ACL)},
  year={2024}
}

@article{deepseekv3,
  title={{DeepSeek-V3} Technical Report},
  author={{DeepSeek-AI}},
  journal={arXiv preprint arXiv:2412.19437},
  year={2024}
}

@article{qwen3moe,
  title={{Qwen3} Technical Report},
  author={{Qwen Team}},
  journal={arXiv preprint arXiv:2505.09388},
  year={2025}
}

@article{wang2024lossfree,
  title={Auxiliary-Loss-Free Load Balancing Strategy for Mixture-of-Experts},
  author={Wang, Lean and others},
  journal={arXiv preprint arXiv:2408.15664},
  year={2024}
}

@article{glmteam2026glm5,
  title={{GLM-5}: From Vibe Coding to Agentic Engineering},
  author={{GLM-5 Team}},
  journal={arXiv preprint arXiv:2602.15763},
  year={2026}
}

@article{kimiteam2025kimik2,
  title={{Kimi K2}: Open Agentic Intelligence},
  author={{Kimi Team}},
  journal={arXiv preprint arXiv:2507.20534},
  year={2025}
}

@article{kimiteam2026kimik3,
  title={{Kimi K3}: Open Frontier Intelligence},
  author={{Kimi Team}},
  journal={arXiv preprint arXiv:2607.24653},
  year={2026}
}

@article{elango2026latentmoe,
  title={{LatentMoE}: Toward Optimal Accuracy per {FLOP} and Parameter in Mixture of Experts},
  author={Elango, Venmugil and Bhatia, Nidhi and Waleffe, Roger and Shafipour, Rasoul and Asida, Tomer and Khattar, Abhinav and Assaf, Nave and Golub, Maximilian and Guman, Joey and Mitra, Tiyasa and Zhao, Ritchie and Borkar, Ritika and Zilberstein, Ran and Patwary, Mostofa and Shoeybi, Mohammad and Rouhani, Bita},
  journal={arXiv preprint arXiv:2601.18089},
  year={2026}
}

@article{cui2026mhlatentmoe,
  title={{Multi-Head} {LatentMoE} and {Head Parallel}: Communication-Efficient and Deterministic {MoE} Parallelism},
  author={Cui, Chenwei and Jackson, Rockwell and Herrera, Benjamin Joseph and T{\'a}rano, Ana Mar{\'i}a and Kerner, Hannah},
  journal={arXiv preprint arXiv:2602.04870},
  year={2026}
}

@techreport{oe2025,
  title={Scaling Embeddings Outperforms Scaling Experts in Language Models},
  author={Liu, Hong and Zhang, Jiaqi and Wang, Chao and Hu, Xing and Lyu, Linkun and Sun, Jiaqi and Yang, Xurui and Wang, Bo and Li, Fengcun and Qian, Yulei and Si, Lingtong and Sun, Yerui and Li, Rumei and Pei, Peng and Xie, Yuchen and Cai, Xunliang},
  institution={Meituan LongCat Team},
  year={2026},
  note={arXiv:2601.21204}
}

@article{team2025longcat,
  title={{LongCat-Flash} Technical Report},
  author={{Meituan LongCat Team}},
  journal={arXiv preprint arXiv:2509.01322},
  year={2025}
}

@inproceedings{hendrycks2021mmlu,
  title={Measuring Massive Multitask Language Understanding},
  author={Hendrycks, Dan and Burns, Collin and Basart, Steven and Zou, Andy and Mazeika, Mantas and Song, Dawn and Steinhardt, Jacob},
  booktitle={International Conference on Learning Representations (ICLR)},
  year={2021}
}

@inproceedings{zellers2019hellaswag,
  title={{HellaSwag}: Can a machine really finish your sentence?},
  author={Zellers, Rowan and Holtzman, Ari and Bisk, Yonatan and Farhadi, Ali and Choi, Yejin},
  booktitle={Proceedings of the 57th Annual Meeting of the Association for Computational Linguistics},
  pages={4791--4800},
  year={2019}
}

@article{wang2024mmlupro,
  title={{MMLU-Pro}: A More Robust and Challenging Multi-Task Language Understanding Benchmark},
  author={Wang, Yubo and Ma, Xueguang and Zhang, Ge and Ni, Yuansheng and Chandra, Abhranil and Guo, Shiguang and Ren, Weiming and Arulraj, Aaran and He, Xuan and Jiang, Ziyan and Li, Tianle and Ku, Max and Wang, Kai and Zhuang, Alex and Fan, Rongqi and Yue, Xiang and Chen, Wenhu},
  journal={arXiv preprint arXiv:2406.01574},
  year={2024}
}

@inproceedings{huang2023ceval,
  title={{C-Eval}: A Multi-Level Multi-Discipline {Chinese} Evaluation Suite for Foundation Models},
  author={Huang, Yuzhen and Bai, Yuzhuo and Zhu, Zhihao and Zhang, Junlei and Zhang, Jinghan and Su, Tangjun and Liu, Junteng and Lv, Chuancheng and Zhang, Yikai and Lei, Jiayi and Fu, Yao and Sun, Maosong and He, Junxian},
  booktitle={Advances in Neural Information Processing Systems (NeurIPS)},
  year={2023}
}

@article{li2023cmmlu,
  title={{CMMLU}: Measuring Massive Multitask Language Understanding in {Chinese}},
  author={Li, Haonan and Zhang, Yixuan and Koto, Fajri and Yang, Yifei and Zhao, Hai and Gong, Yeyun and Duan, Nan and Baldwin, Timothy},
  journal={arXiv preprint arXiv:2306.09212},
  year={2023}
}

@inproceedings{suzgun2023bbh,
  title={Challenging {BIG-Bench} Tasks and Whether Chain-of-Thought Can Solve Them},
  author={Suzgun, Mirac and Scales, Nathan and Sch{\"a}rli, Nathanael and Gehrmann, Sebastian and Tay, Yi and Chung, Hyung Won and Chowdhery, Aakanksha and Le, Quoc V. and Chi, Ed H. and Zhou, Denny and Wei, Jason},
  booktitle={Findings of the Association for Computational Linguistics (ACL Findings)},
  year={2023}
}

@inproceedings{rein2024gpqa,
  title={{GPQA}: A Graduate-Level {Google}-Proof {Q\&A} Benchmark},
  author={Rein, David and Hou, Betty Li and Stickland, Asa Cooper and Petty, Jackson and Pang, Richard Yuanzhe and Dirani, Julien and Michael, Julian and Bowman, Samuel R.},
  booktitle={Conference on Language Modeling (COLM)},
  year={2024}
}

@article{cobbe2021gsm8k,
  title={Training Verifiers to Solve Math Word Problems},
  author={Cobbe, Karl and Kosaraju, Vineet and Bavarian, Mohammad and Chen, Mark and Jun, Heewoo and Kaiser, Lukasz and Plappert, Matthias and Tworek, Jerry and Hilton, Jacob and Nakano, Reiichiro and Hesse, Christopher and Schulman, John},
  journal={arXiv preprint arXiv:2110.14168},
  year={2021}
}

@article{shi2022language,
  title={Language models are multilingual chain-of-thought reasoners},
  author={Shi, Freda and Suzgun, Mirac and Freitag, Markus and Wang, Xuezhi and Srivats, Suraj and Vosoughi, Soroush and Chung, Hyung Won and Tay, Yi and Ruder, Sebastian and Zhou, Denny and others},
  journal={arXiv preprint arXiv:2210.03057},
  year={2022}
}

@article{wei2023cmath,
  title={{CMATH}: Can Your Language Model Pass {Chinese} Elementary School Math Test?},
  author={Wei, Tianwen and Luan, Jian and Liu, Wei and Dong, Shuang and Wang, Bin},
  journal={arXiv preprint arXiv:2306.16636},
  year={2023}
}

@inproceedings{liu2023your,
  title={Is your code generated by {ChatGPT} really correct? {R}igorous evaluation of large language models for code generation},
  author={Liu, Jiawei and Xia, Chunqiu Steven and Wang, Yuyao and Zhang, Lingming},
  booktitle={Advances in Neural Information Processing Systems (NeurIPS)},
  volume={36},
  pages={21558--21572},
  year={2023}
}

@article{hendrycks2021measuring,
  title={Measuring mathematical problem solving with the {MATH} dataset},
  author={Hendrycks, Dan and Burns, Collin and Kadavath, Saurav and Arora, Akul and Basart, Steven and Tang, Eric and Song, Dawn and Steinhardt, Jacob},
  journal={arXiv preprint arXiv:2103.03874},
  year={2021}
}

@inproceedings{kornblith2019similarity,
  title={Similarity of Neural Network Representations Revisited},
  author={Kornblith, Simon and Norouzi, Mohammad and Lee, Honglak and Hinton, Geoffrey},
  booktitle={Proceedings of the 36th International Conference on Machine Learning},
  volume={97},
  pages={3519--3529},
  year={2019},
  publisher={PMLR},
  url={https://proceedings.mlr.press/v97/kornblith19a.html}
}

@misc{deepgemm2025,
  title={{DeepGEMM}: Clean and Efficient {BLAS} Kernel Library on {GPU}},
  author={Zhao, Chenggang and Xu, Zhean and Zhao, Liang and Li, Jiashi and Xu, Chenhao and Xu, Anyi and Liu, Shengyu and Zhou, Kexing and Yu, Kuai},
  year={2025},
  publisher={GitHub},
  howpublished={\url{https://github.com/deepseek-ai/DeepGEMM}}
}

@inproceedings{yang2021tensor,
  title={Tensor Programs {IV}: Feature Learning in Infinite-Width Neural Networks},
  author={Yang, Greg and Hu, Edward J.},
  booktitle={Proceedings of the 38th International Conference on Machine Learning},
  series={Proceedings of Machine Learning Research},
  volume={139},
  pages={11727--11737},
  year={2021},
  publisher={PMLR},
  url={https://proceedings.mlr.press/v139/yang21c.html}
}

@inproceedings{everett2024scaling,
  title={Scaling Exponents Across Parameterizations and Optimizers},
  author={Everett, Katie E. and Xiao, Lechao and Wortsman, Mitchell and Alemi, Alexander A. and Novak, Roman and Liu, Peter J. and Gur, Izzeddin and Sohl-Dickstein, Jascha and Kaelbling, Leslie Pack and Lee, Jaehoon and Pennington, Jeffrey},
  booktitle={Proceedings of the 41st International Conference on Machine Learning},
  series={Proceedings of Machine Learning Research},
  volume={235},
  pages={12666--12700},
  year={2024},
  publisher={PMLR},
  url={https://proceedings.mlr.press/v235/everett24a.html}
}

\appendix
\section{Update Alignment and the Choice of Forward Scaling}\label{app:update-alignment}

\paragraph{Initialization-matched reference.}
Both initialization-only scaling $(0, 1/2, 0)$ and forward scaling $(1/2, 0, 0)$ satisfy $a + b = 1/2$ and restore the step-zero scale.
To isolate how forward scaling compensates for parameter-induced updates during training, we compare \gmoe{} against the initialization-only variant under the setup of Section~\ref{sec:analysis:compatibility}, using the latter as a reference for diagnosing update accumulation.

\paragraph{Update alignment.}
Adapting the finite-size log-alignment diagnostic of \citet{everett2024scaling} to a single-step parameter update $\Delta W$, we compute
\begin{equation}
\widehat{\alpha}
=\log_{\dsub}\!\left(
\frac{\mathrm{RMS}(\Delta Wz)}
{\mathrm{RMS}(\Delta W)\,\mathrm{RMS}(z)}
\right),
\label{eq:empirical-update-alignment}
\end{equation}
where $z$ is held fixed at its pre-update value. We measure routers and sampled expert gate/up projections across layers and groups.

\begin{figure}[htbp]
\centering
\includegraphics[width=\linewidth]{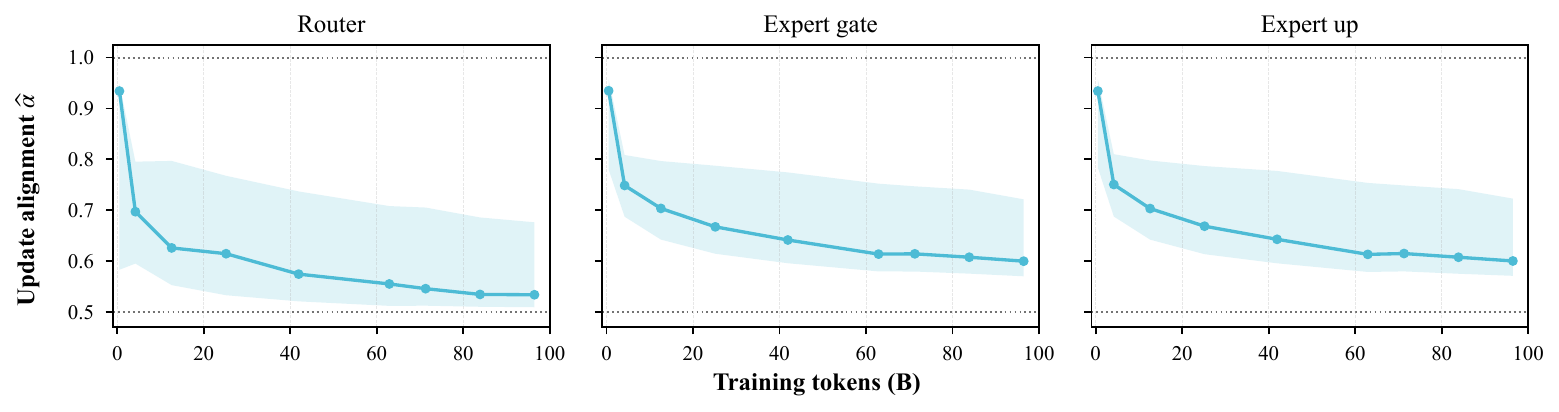}
\caption{Update alignment along the initialization-only trajectory. Lines show medians and shaded bands show the 10th to 90th percentiles across layers, groups, and sampled experts. Dotted lines mark $\widehat{\alpha}=1/2$ and $1$.}
\label{fig:update-alignment}
\end{figure}

\begin{figure}[!htbp]
\centering
\includegraphics[width=0.60\linewidth]{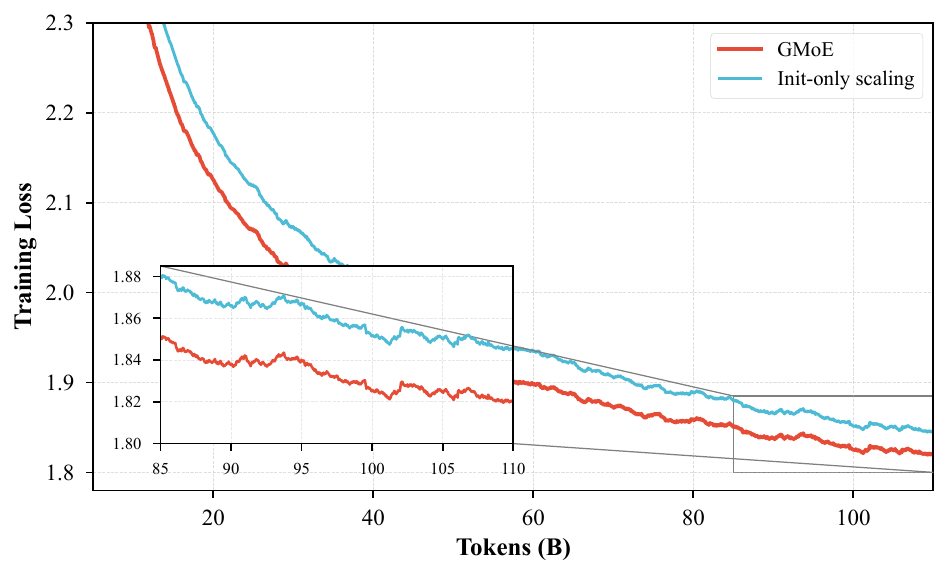}
\caption{Forward scaling versus initialization-only scaling. \gmoe achieves lower training loss throughout the displayed interval. Curves are EMA-smoothed, and the inset enlarges the final training phase.}
\label{fig:init-only-loss}
\end{figure}

\paragraph{Implications for forward scaling.}
Within the scaling model of Eq.~\eqref{eq:update_ratio}, matching the direct update scale at an unchanged learning rate ($c=0$) requires $a=\alpha$. The fully aligned case $\alpha=1$ therefore suggests a forward multiplier $\ngroup$, paired with $b=-1/2$ to preserve the initial activation scale. Figure~\ref{fig:update-alignment} shows that the measured alignment shifts from near full alignment toward the square-root accumulation regime as training progresses. In the same scaling model, $\sqrt{\ngroup}$ forward scaling yields $\ngroup^{1/2-\alpha}$, whereas $\ngroup$ forward scaling yields $\ngroup^{1-\alpha}$ and amplifies the direct update relative to the baseline when $\alpha<1$. Together with the loss comparison in Figure~\ref{fig:init-only-loss}, the observed partial alignment supports $(a,b,c)=(1/2,0,0)$, which retains the baseline initialization and learning rate while compensating for the attenuated updates of the initialization-only reference.

\paragraph{Training performance.}
Figure~\ref{fig:init-only-loss} compares the two parameterizations through $110$B training tokens. Initialization-only scaling consistently yields higher training loss than \gmoe with forward scaling. At the same evaluation step as Table~\ref{tab:align-ablation}, its training and validation losses are $1.827$ and $2.970$, compared with $1.802$ and $2.940$ for \gmoe (Table~\ref{tab:init-only-loss}). Matching the initial activation scale alone therefore does not recover the training performance of forward scaling.

\begin{table}[!htbp]
\centering
\small
\setlength{\tabcolsep}{8pt}
\renewcommand{\arraystretch}{1.08}
\caption{Comparison with initialization-only scaling.}
\label{tab:init-only-loss}
\begin{tabular}{lcc}
\toprule
Variant & Training loss $\downarrow$ & Validation loss $\downarrow$ \\
\midrule
\gmoe & \textbf{1.802} & \textbf{2.940} \\
Init-only scaling & 1.827 & 2.970 \\
\bottomrule
\end{tabular}
\end{table}

\FloatBarrier

\section{Cross-Group Similarity and Routing Neighbors}\label{app:group-analysis}

\paragraph{Linear CKA.}\label{app:group-cka}
We measure cross-group similarity using linear centered kernel alignment (CKA)~\citep{kornblith2019similarity}. Let $X\in\mathbb{R}^{T\times p}$ and $Y\in\mathbb{R}^{T\times q}$ contain the representations of the same $T$ tokens from two groups. After subtracting the mean of each feature across tokens, we obtain $\widetilde X$ and $\widetilde Y$ and compute
\begin{equation}\label{eq:group-cka}
    \operatorname{CKA}(X,Y) =
    \frac{\lVert\widetilde X^\top\widetilde Y\rVert_F^2}
    {\lVert\widetilde X^\top\widetilde X\rVert_F\,\lVert\widetilde Y^\top\widetilde Y\rVert_F},
\end{equation}
where $\lVert\cdot\rVert_F$ denotes the Frobenius norm.
We apply this measure to group inputs, routing weights, and group outputs projected back to the shared residual space. For routing weights, each row is a sparse top-$\kexp$ vector with zero entries for unselected experts, and selected weights normalized to sum to one. Each comparison uses $50{,}000$ validation-set tokens. We average CKA over the six unordered pairs of the four groups within each layer, then average the layer-level values to obtain the results in Table~\ref{tab:group-cka-stages}.

\paragraph{Routing neighbors.}\label{app:routing_neighbors}

To identify the routing neighbors of an anchor token $a$ within a specific group $g$ (as presented in Table~\ref{tab:routing-examples}), we first compute its group-specific routing prototype. Let $r_g(x_i)$ denote the complete routing weight vector for the $i$-th occurrence of the token in the corpus. We average the $L_1$-normalized routing weights over all $N_a$ occurrences of the token and apply $L_2$ normalization to obtain the prototype $p_g(a)$:
\begin{equation}
    \bar{r}_g(a) = \frac{1}{N_a} \sum_{i:\,x_i=a} \frac{r_g(x_i)}{\lVert r_g(x_i)\rVert_1}, \qquad p_g(a) = \frac{\bar{r}_g(a)}{\lVert \bar{r}_g(a)\rVert_2}.
\end{equation}
Routing similarity between anchor $a$ and candidate $b$ is the cosine similarity of their prototypes:
\begin{equation}
    s_g(a,b) = p_g(a)^\top p_g(b).
\end{equation}
Finally, the routing neighbors of $a$ in group $g$ are defined as the top-$k$ tokens that maximize this similarity:
\begin{equation}
    \mathcal{N}_g^k(a) = \underset{b \neq a}{\operatorname{TopK}}\, s_g(a,b).
\end{equation}
Intuitively, these neighbors are not necessarily words with high embedding cosine similarity, nor are they simply frequently co-occurring words in the text. Rather, they are tokens that consistently trigger similar expert-selection patterns within a specific group's representational subspace.

\section{Distributed Execution of a GMoE Layer}\label{app:distributed_execution}

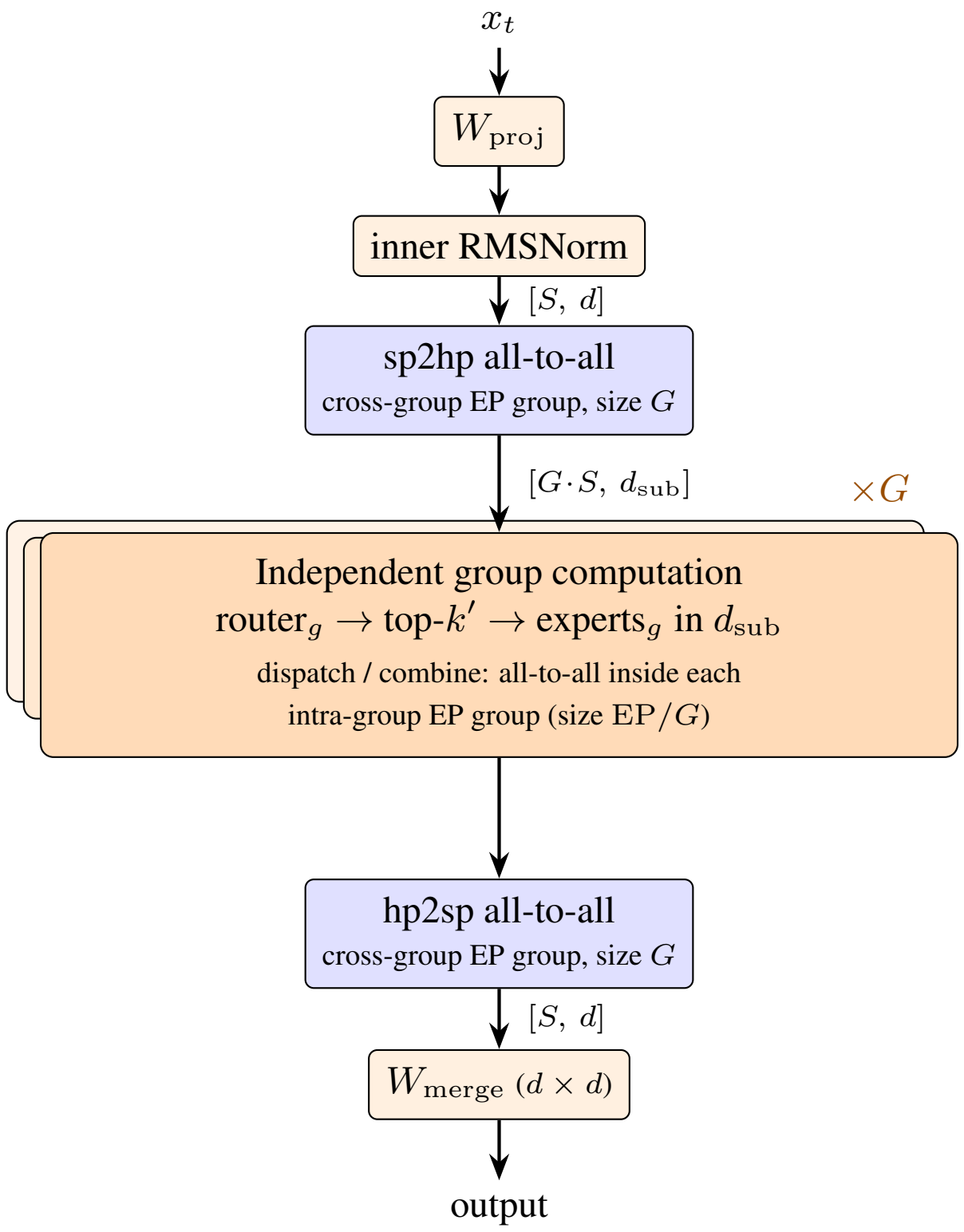
\begin{figure}[H]
\centering
\begin{adjustbox}{max width=\linewidth}
\begin{tikzpicture}[
    font=\small,
    comp/.style={draw, rounded corners=2pt, align=center, inner sep=4pt, fill=orange!12},
    comm/.style={draw, rounded corners=2pt, align=center, inner sep=4pt, fill=blue!12},
    grp/.style={draw, rounded corners=3pt, align=center, inner sep=6pt, fill=orange!28, minimum width=7.6cm, minimum height=1.5cm},
    arr/.style={-{Stealth[length=2.2mm]}, thick},
  ]
  \node[] (input) {$x_t$};
  \node[comp, below=4mm of input] (proj) {$W_{\mathrm{proj}}$};
  \node[comp, below=4mm of proj] (norm) {inner RMSNorm};
  \node[comm, below=4mm of norm] (sp2hp) {sp2hp all-to-all\\ \scriptsize cross-group EP group, size $\ngroup$};
  \draw[arr] (input) -- (proj);
  \draw[arr] (proj) -- (norm);
  \draw[arr] (norm) -- node[midway, right=1mm, font=\scriptsize] {$[S,\,d]$} (sp2hp);

  \node[grp, below=7mm of sp2hp, xshift=-2.8mm, fill=orange!10] (gback2) {};
  \node[grp, fill=orange!18] at ($(gback2.center)+(1.4mm,-1.4mm)$) (gback1) {};
  \node[grp] at ($(gback1.center)+(1.4mm,-1.4mm)$) (groups) {Independent group computation\\[1pt] router$_g$ $\to$ top-$\kexp$ $\to$ experts$_g$ in $\dsub$\\[2pt] \scriptsize dispatch / combine: all-to-all inside each\\ \scriptsize intra-group EP group (size $\mathrm{EP}/\ngroup$)};
  \draw[arr] (sp2hp.south) -- node[midway, right=1mm, font=\scriptsize] {$[\ngroup\!\cdot\!S,\,\dsub]$} (groups.north);
  \node[anchor=south east, font=\footnotesize\bfseries, text=orange!60!black] at (gback2.north east) {$\times\ngroup$};

  \node[comm, below=10mm of groups] (hp2sp) {hp2sp all-to-all\\ \scriptsize cross-group EP group, size $\ngroup$};
  \draw[arr] (groups.south) -- (hp2sp.north);

  \node[comp, below=5mm of hp2sp] (merge) {$W_{\mathrm{merge}}$ \scriptsize ($d\times d$)};
  \node[below=5mm of merge] (output) {output};
  \draw[arr] (hp2sp) -- node[midway, right=1mm, font=\scriptsize] {$[S,\,d]$} (merge);
  \draw[arr] (merge) -- (output);

\end{tikzpicture}
\end{adjustbox}
\caption{Data flow of a \gmoe layer (Section~\ref{sec:system:comm}, notation as in the text). The subspace split and its inverse are each realized by one all-to-all over the cross-group expert-parallel group (blue). Routing and expert computation proceed independently within each subspace (orange), including the intra-group dispatch/combine all-to-alls.}
\label{fig:gmoe-comm}
\end{figure}

Figure~\ref{fig:gmoe-comm} summarizes the distributed execution described in Section~\ref{sec:system:comm}, including the cross-group layout conversions and the routing and expert computation within each group.

\paragraph{Layout conversion.}
Before sp2hp, each rank holds $S$ tokens with full $d$-dimensional features. The normalized projection is split into $\ngroup$ feature chunks, and a cross-group all-to-all delivers each chunk to its assigned rank. Each rank then holds $\ngroup S$ tokens of width $\dsub=d/\ngroup$. After local or intra-group expert execution, hp2sp reverses this exchange and restores the $[S,d]$ layout before the merge projection.

\begin{figure}[!htbp]
\centering
\begin{subfigure}{\textwidth}
\centering
\begin{tikzpicture}[font=\scriptsize,
    rank/.style={draw, rounded corners=1pt, minimum width=14mm, minimum height=7mm, inner sep=1pt, align=center},
    src/.style={rank, fill=green!25},
    arr/.style={-{Stealth[length=1.8mm]}, thick},
  ]
  \node[src] (r0) {r0\\[0pt][dp0,cp0]};
  \node[rank, right=1.2mm of r0] (r1) {r1\\[0pt][dp0,cp1]};
  \node[src, right=1.2mm of r1] (r2) {r2\\[0pt][dp0,cp2]};
  \node[rank, right=1.2mm of r2] (r3) {r3\\[0pt][dp0,cp3]};
  \node[src, right=1.2mm of r3] (r4) {r4\\[0pt][dp1,cp0]};
  \node[rank, right=1.2mm of r4] (r5) {r5\\[0pt][dp1,cp1]};
  \node[src, right=1.2mm of r5] (r6) {r6\\[0pt][dp1,cp2]};
  \node[rank, right=1.2mm of r6] (r7) {r7\\[0pt][dp1,cp3]};
  \node[draw, fill=blue!8, align=center, below=13mm of r3, inner sep=4pt, minimum width=10cm] (dst) {r0 after sp2hp:\\[1pt] $[\mathrm{dp0,cp0}\,|\,\mathrm{dp0,cp2}\,|\,\mathrm{dp1,cp0}\,|\,\mathrm{dp1,cp2}]\otimes d_0$};
  \draw[arr] (r0.south) -- (r0.south |- dst.north);
  \draw[arr] (r2.south) -- (r2.south |- dst.north);
  \draw[arr] (r4.south) -- (r4.south |- dst.north);
  \draw[arr] (r6.south) -- (r6.south |- dst.north);
\end{tikzpicture}
\caption{$\mathrm{CP}=4$, $w=\mathrm{EP}/\ngroup=2$ ($\mathrm{DP}{=}2$, $\mathrm{EP}{=}8$, $\ngroup=4$): the stride steps along the CP axis and wraps into the next DP row. Partners occupy cp0 and cp2 across both DP rows.}
\label{fig:sp2hp-cp-gt}
\end{subfigure}

\vspace{2mm}
\begin{subfigure}{\textwidth}
\centering
\begin{tikzpicture}[font=\scriptsize,
    rank/.style={draw, rounded corners=1pt, minimum width=14mm, minimum height=7mm, inner sep=1pt, align=center},
    src/.style={rank, fill=green!25},
    arr/.style={-{Stealth[length=1.8mm]}, thick},
  ]
  \node[src] (r0) {r0\\[0pt][dp0,cp0]};
  \node[rank, right=1.2mm of r0] (r1) {r1\\[0pt][dp0,cp1]};
  \node[rank, right=1.2mm of r1] (r2) {r2\\[0pt][dp1,cp0]};
  \node[rank, right=1.2mm of r2] (r3) {r3\\[0pt][dp1,cp1]};
  \node[src, right=1.2mm of r3] (r4) {r4\\[0pt][dp2,cp0]};
  \node[rank, right=1.2mm of r4] (r5) {r5\\[0pt][dp2,cp1]};
  \node[rank, right=1.2mm of r5] (r6) {r6\\[0pt][dp3,cp0]};
  \node[rank, right=1.2mm of r6] (r7) {r7\\[0pt][dp3,cp1]};
  \node[draw, fill=blue!8, align=center, below=13mm of r2, inner sep=4pt, minimum width=7cm] (dst) {r0 after sp2hp:\\[1pt] $[\mathrm{dp0,cp0}\,|\,\mathrm{dp2,cp0}]\otimes d_0$};
  \draw[arr] (r0.south) -- (r0.south |- dst.north);
  \draw[arr] (r4.south) -- (r4.south |- dst.north);
\end{tikzpicture}
\caption{$\mathrm{CP}=2$ divides $w=\mathrm{EP}/\ngroup=4$ ($\mathrm{DP}{=}4$, $\mathrm{EP}{=}8$, $\ngroup=2$): the stride walks the DP axis. Partners share cp0 but differ in dp.}
\label{fig:sp2hp-cp-lt}
\end{subfigure}
\caption{Where the sp2hp all-to-all draws tokens from. The EP domain occupies $\mathrm{EP}/\mathrm{CP}$ complete DP rows of the (DP, CP) grid, numbered DP-major (the CP index varies fastest). Intra-group expert-parallel groups are $w=\mathrm{EP}/\ngroup$ consecutive ranks, so the cross-group expert-parallel group (size $\ngroup$) strides by exactly $w$. Every rank initially holds all $\ngroup$ hidden slices ($d_{0..\ngroup-1}$) of its own (dp, cp) shard. Green ranks contribute tokens to rank 0's group-0 slice after the exchange (i.e., hidden slice $d_0$). Here $[\,\cdots\,]\otimes d_0$ concatenates the listed shards' $d_0$ slices (Section~\ref{sec:system:comm}).}
\label{fig:sp2hp-cp-mapping}
\end{figure}
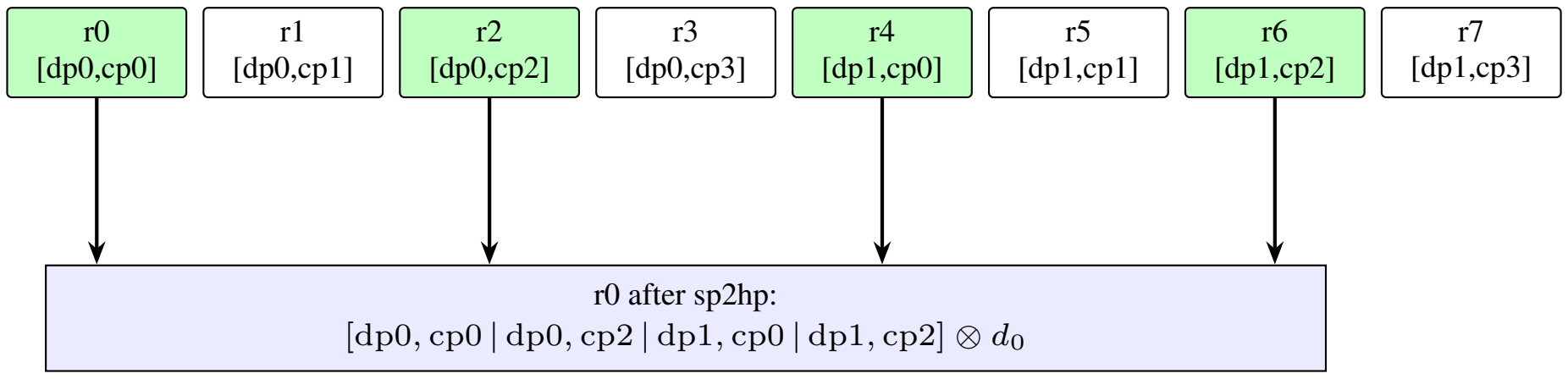
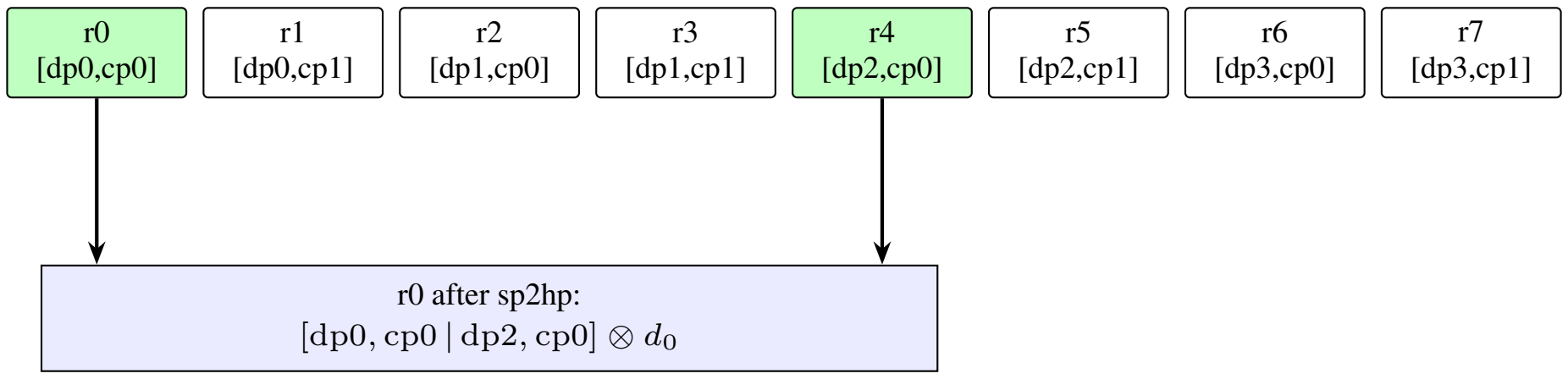

\paragraph{Device-grid mapping.}
Which ranks exchange tokens in sp2hp depends on how the EP domain is laid out on the (DP, CP) device grid.
In our implementation $\mathrm{TP}=1$ and CP divides $\mathrm{EP}$, so an EP domain always spans $\mathrm{EP}/\mathrm{CP}$ consecutive DP rows, numbered DP-major (the CP index varies fastest).
Let $r\in\{0,\ldots,\mathrm{EP}-1\}$ denote a rank's local index within the EP domain, and let $w=\mathrm{EP}/\ngroup$ be the stride between consecutive members of a cross-group EP group.
The members of rank $r$'s cross-group EP group have local indices $r_j=(r\bmod w)+jw$ for $j=0,\ldots,\ngroup-1$.
Their grid coordinates within the EP domain are $(\mathrm{dp},\mathrm{cp})=\bigl(\lfloor r_j/\mathrm{CP}\rfloor,\ r_j\bmod\mathrm{CP}\bigr)$.
Two patterns matter in practice.
If $w<\mathrm{CP}$, the stride proceeds along the CP axis and wraps into the next DP row: with $\mathrm{DP}{=}2$, $\mathrm{CP}{=}4$, $\mathrm{EP}{=}8$, $\ngroup{=}4$ ($w{=}2$), rank 0's group-0 slice gathers tokens from (dp0,cp0), (dp0,cp2), (dp1,cp0) and (dp1,cp2) (Figure~\ref{fig:sp2hp-cp-gt}).
If CP divides $w$, the stride proceeds along the DP axis at a fixed CP position: with $\mathrm{DP}{=}4$, $\mathrm{CP}{=}2$, $\mathrm{EP}{=}8$, $\ngroup{=}2$ ($w{=}4$), rank 0 gathers from (dp0,cp0) and (dp2,cp0) (Figure~\ref{fig:sp2hp-cp-lt}).
The physical locality (intra- vs.\ inter-node) of the sp2hp/hp2sp traffic changes with the layout, and its cost changes with it.

\clearpage
\section{Communication Volume Accounting}\label{app:comm_accounting}

We derive the per-rank token communication volume for the two expert layouts evaluated in Section~\ref{sec:system:comm} and compare the theoretical values with hardware counters.

\paragraph{Equal-partition exchanges.}
Each sp2hp or hp2sp call inputs $Sdb$ bytes per rank, where $b$ is the element width.
Of these, a $1/P$ fraction stays local when the all-to-all spans $P$ ranks.
The cross-rank application-level send volume is therefore
\begin{equation}\label{eq:comm-off}
  B_{\mathrm{off}} = B_{\mathrm{API}}\!\left(1-\frac{1}{P}\right).
\end{equation}

\paragraph{Variable-length expert dispatch.}
For a dispatch or combine call whose $c$-th invocation sends rows of width $a_c$ at $b_c$ bytes per element, with $s_{c,u}$ rows destined for member $u$ and self-index $i_c$:
\begin{equation}\label{eq:comm-varlen}
  B_{\mathrm{API},c}=b_c a_c\!\sum_u s_{c,u},\qquad
  B_{\mathrm{off},c}=b_c a_c\!\sum_{u\ne i_c} s_{c,u}.
\end{equation}

\paragraph{Local-group layout ($\mathrm{EP}=\ngroup=4$).}
The four layout exchanges (sp2hp and hp2sp in the forward and backward passes) account for the module's cross-rank token communication.
With $S=8192$, $d=3072$, $b=2$ (BF16):
\begin{equation}\label{eq:comm-ep4}
  B_{\mathrm{API,FB}}^{\mathrm{EP}=4}=4Sdb=192\;\mathrm{MiB},\qquad
  B_{\mathrm{off,FB}}^{\mathrm{EP}=4}=4Sdb\!\left(1-\tfrac{1}{4}\right)=144\;\mathrm{MiB}.
\end{equation}

\paragraph{Split layout ($\mathrm{EP}=8$, each group on two ranks).}
Beyond the four layout exchanges, the forward and backward passes add four BF16 expert-vector exchanges, two FP32 exchanges carrying routing probabilities and their gradients, and one int64 expert-count exchange.
Let $\bar{n}$ denote the mean per-rank dispatch row count and $\bar{r}$ the mean rows sent to the other rank, both from actual routing decisions ($\bar{n}=266149.125$, $\bar{r}=133654.125$).
Let $b_x=2$, $b_p=4$, and $b_m=8$ denote the bytes per element for expert vectors and their gradients, routing probabilities and their gradients, and expert counts, respectively. With $N_{\mathrm{F}}=256$ FFN experts per group, we obtain:
\begin{equation}\label{eq:comm-ep8}
\begin{aligned}
  B_{\mathrm{API,FB}}^{\mathrm{EP}=8}
    &= 4Sdb_x + 4\bar{n}\tfrac{d}{\ngroup}b_x + 2\bar{n}\,b_p + N_{\mathrm{F}}b_m
    = 1753.50\;\mathrm{MiB},\\
  B_{\mathrm{off,FB}}^{\mathrm{EP}=8}
    &= 4Sdb_x\!\left(1{-}\tfrac{1}{4}\right) + 4\bar{r}\tfrac{d}{\ngroup}b_x + 2\bar{r}\,b_p + \tfrac{N_{\mathrm{F}}b_m}{2}
    = 928.15\;\mathrm{MiB}.
\end{aligned}
\end{equation}

\begin{table}[H]
\centering
\small
\setlength{\tabcolsep}{5pt}
\caption{Token communication volume per GPU over the forward and backward passes for the two expert layouts in Section~\ref{sec:system:comm}. Theoretical cross-rank volumes are computed from Eqs.~\eqref{eq:comm-ep4}--\eqref{eq:comm-ep8}. Measured values are NVLink send counters averaged over all GPUs.}
\label{tab:comm-accounting}
\begin{tabular}{@{}lrrrr@{}}
\toprule
Layout & \makecell{API input\\(MiB)} & \makecell{Theoretical\\cross-rank (MiB)} & \makecell{Measured\\NVLink (MiB)} & \makecell{Relative\\deviation} \\
\midrule
Local-group (EP4/EDP2) & 192.000 & 144.000 & 144.021 & ${+}0.015\%$ \\
Split (EP8/EDP1)          & 1753.500 & 928.150 & 928.273 & ${+}0.013\%$ \\
\bottomrule
\end{tabular}
\end{table}

Table~\ref{tab:comm-accounting} itemizes the theoretical and measured volumes.
The local-group layout reduces cross-rank token communication to four fixed-size layout exchanges, lowering the send volume from $928$ to $144$\,MiB per GPU.
The split layout additionally exchanges expert vectors, routing probabilities, their gradients, and expert counts across ranks, with volumes that depend on the actual routing decisions.

\clearpage
\section{Detailed Model Specifications and Hyperparameters}\label{app:training_config}

The models in Table~\ref{tab:app-model-config} build on the LongCat-Flash architecture~\citep{team2025longcat}, following the LongCat-Flash-Lite backbone with $N$-gram embeddings~\citep{oe2025}. Shared-expert FFN blocks and MoE blocks are interleaved in a shortcut-connected MoE (ScMoE) layout. Each shortcut layer contains two attention blocks, with the first attention output connected directly to the MoE module. Zero-computation experts act as identity mappings and participate in routing alongside FFN experts.

\begin{table}[!htbp]
\centering
\small
\setlength{\tabcolsep}{5pt}
\renewcommand{\arraystretch}{1.08}
\caption{Model specifications and training hyperparameters for the large-scale experiments.}
\label{tab:app-model-config}
\begin{tabular}{@{}p{0.40\linewidth}*{3}{>{\centering\arraybackslash}p{\dimexpr0.20\linewidth-2\tabcolsep\relax}}@{}}
\toprule
Attribute & Fine-grained MoE & \gmoe & LatentMoE \\
\midrule
\multicolumn{4}{@{}l}{\textbf{Parameter counts}} \\
Total parameters (B) & 68.8 & 69.0 & 68.9 \\
Activated parameters (B)$^{*}$ & 3.6 & 3.8 & 3.7 \\
$N$-gram embedding parameters (B) & \multicolumn{3}{c@{}}{31.4} \\
\midrule
\multicolumn{4}{@{}l}{\textbf{Model architecture}} \\
Shortcut layers & \multicolumn{3}{c@{}}{14} \\
Hidden dimension & \multicolumn{3}{c@{}}{3072} \\
Attention variant & \multicolumn{3}{c@{}}{MLA} \\
Attention heads & \multicolumn{3}{c@{}}{32} \\
Query compression rank & \multicolumn{3}{c@{}}{1536} \\
KV compression rank & \multicolumn{3}{c@{}}{512} \\
QK head dimension, non-RoPE & \multicolumn{3}{c@{}}{128} \\
QK head dimension, RoPE & \multicolumn{3}{c@{}}{64} \\
Value head dimension & \multicolumn{3}{c@{}}{128} \\
Shared-expert intermediate dimension & \multicolumn{3}{c@{}}{6144} \\
\midrule
\multicolumn{4}{@{}l}{\textbf{MoE configuration}} \\
Routing groups & 1 & 4 & 1 \\
Expert input dimension & 3072 & 768 & 768 \\
Router input dimension & 3072 & 768 & 3072 \\
FFN experts per module & 256 & $4\times256$ & 1024 \\
Zero experts per module & 128 & $4\times128$ & 512 \\
Expert selections per token & 12 & $4\times12$ & 48 \\
Expert intermediate dimension & \multicolumn{3}{c@{}}{1024} \\
Input projection & None & $3072\to3072$ & $3072\to768$ \\
Merge projection & None & $3072\to3072$ & $768\to3072$ \\
Loss-free bias update rate & \multicolumn{3}{c@{}}{0.01} \\
Auxiliary balancing coefficient & \multicolumn{3}{c@{}}{0} \\
\midrule
\multicolumn{4}{@{}l}{\textbf{Training hyperparameters}} \\
Sequence length & \multicolumn{3}{c@{}}{8192} \\
Optimizer & \multicolumn{3}{c@{}}{AdamW} \\
AdamW $\beta_1,\beta_2$ & \multicolumn{3}{c@{}}{$(0.9,0.95)$} \\
AdamW $\epsilon$ & \multicolumn{3}{c@{}}{$10^{-16}$} \\
Base learning rate & \multicolumn{3}{c@{}}{$4.5\times10^{-4}$} \\
Learning-rate schedule & \multicolumn{3}{c@{}}{Constant after warmup} \\
Warmup steps & \multicolumn{3}{c@{}}{1000} \\
Global batch size, sequences & \multicolumn{3}{c@{}}{$1024\to2048\to4096\to8192$} \\
Batch-size transition steps & \multicolumn{3}{c@{}}{$12{,}000,\ 32{,}000,\ 72{,}000$} \\
Weight decay & \multicolumn{3}{c@{}}{0.1} \\
Gradient clipping & \multicolumn{3}{c@{}}{1.0} \\
\bottomrule
\end{tabular}

\vspace{4pt}
\begin{minipage}{\linewidth}
\footnotesize
Expert counts and selections are reported per MoE module. Selections include identity zero experts.

$^{*}$Activated parameter counts exclude embedding tables and the output head, using the configured target averages of 8, $4\times8$, and 32 nonzero expert selections per token per MoE module, respectively.
\end{minipage}
\end{table}

\end{document}